\documentclass[sigconf]{acmart}

\usepackage{multirow}
\usepackage{enumitem}
\usepackage{makecell}
\usepackage{subcaption}
\usepackage[dvipsnames,table]{xcolor}
\usepackage{threeparttable}

\usepackage{balance}
\newcommand{\first}[1]{\cellcolor{lightgray!30}{\textbf{#1}}}
\newcommand{\capfirst}[1]{\colorbox{lightgray!30}{\textbf{#1}}}

\newcommand{\capblue}[1]{\colorbox[HTML]{F4F9FD}{\textcolor[HTML]{3E6285}{\textbf{#1}}}}
\newcommand{\capred}[1]{\colorbox[HTML]{FDEDEE}{\textcolor[HTML]{E91739}{\textbf{#1}}}}

\AtBeginDocument{%
  }

\setcopyright{none}
\acmConference[arXiv Preprint]{arXiv Preprint}{2025}{}
\acmBooktitle{}
\acmDOI{}
\acmISBN{}
\acmPrice{}

\begin{document}

\title{Stochastic Deep Graph Clustering for Practical Group Formation}



\author{Junhyung Park}
\orcid{0009-0000-6714-9684}
\affiliation{%
  \institution{Inha University}
  \city{Incheon}
  \country{South Korea}
}
\email{quixote1103@inha.edu}

\author{Hyungjin Kim}
\orcid{0009-0002-3409-2081}
\affiliation{%
  \institution{Inha University}
  \city{Incheon}
  \country{South Korea}
}
\email{flslzk@inha.edu}

\author{Seokho Ahn}
\orcid{0000-0002-5715-4057}
\affiliation{%
  \institution{Inha University}
  \city{Incheon}
  \country{South Korea}
}
\email{sokho0514@inha.edu}

\author{Young-Duk Seo}
\authornote{Corresponding Author}
\orcid{0000-0001-8542-2058}
\affiliation{%
  \institution{Inha University}
  \city{Incheon}
  \country{South Korea}
}
\email{mysid88@inha.ac.kr}


\begin{abstract}
While prior work on group recommender systems (GRSs) has primarily focused on improving recommendation accuracy, most approaches assume static or predefined groups, making them unsuitable for dynamic, real-world scenarios. 
We reframe group formation as a core challenge in GRSs and propose \textsf{\textbf{DeepForm}} (\textsf{Stochastic \textbf{Deep} Graph Clustering for Practical Group \textbf{Form}ation}), a framework designed to meet three key operational requirements: 
(1) the incorporation of high-order user information, 
(2) real-time group formation, and 
(3) dynamic adjustment of the number of groups.
DeepForm employs a lightweight GCN architecture that effectively captures high-order structural signals. 
Stochastic cluster learning enables adaptive group reconfiguration without retraining, while 
contrastive learning refines groups under dynamic conditions.
Experiments on multiple datasets demonstrate that \textsf{\textbf{DeepForm}} achieves superior group formation quality, efficiency, and recommendation accuracy compared with various baselines.
\end{abstract}

\ccsdesc[500]{Information systems~Recommender systems}
\ccsdesc[500]{Information systems~Clustering and classification}

\keywords{Group Recommender Systems, Group Formation, Deep Graph Clustering}


\maketitle

\section{Introduction}
\label{sec:intro}

Humans are inherently social beings who often participate in group-oriented activities, such as dining with colleagues, watching television with family, or attending movies with friends \cite{xu2024aligngroup}.
With the proliferation of online platforms, users increasingly form virtual communities centered around shared interests in domains such as movies \cite{sankar2020groupim}, music \cite{htun2021perception}, and travel \cite{alves2022grouplanner}.  
This growing prevalence of group-based behavior has heightened the importance of group recommender systems (GRSs), which aim to generate recommendations tailored to the collective preferences of user groups \cite{o2001polylens,xu2024aligngroup}.

A central factor in GRSs is group formation, which is the process of organizing users into groups whose preferences can be aggregated for recommendation. Grouping users with similar tastes allows GRSs to better capture group characteristics and produce recommendations that more accurately reflect collective preferences, ultimately enhancing user satisfaction \cite{krouska2023novel}. For effective and efficient group formation in practical settings, three operational requirements must be satisfied (Section \ref{subsec:requirements}): (1) Incorporation of high-order graph structures, (2) Real-time group formation, and (3) Dynamic determination of the optimal number of groups ($K$).

Despite the importance of these requirements, most existing GRS studies sidestep group formation, relying instead on predefined groups \cite{sankar2020groupim,chen2022thinking,wu2023consrec} or simple similarity heuristics \cite{ceh2022performance,stratigi2022sequential}. 
Such static approaches cannot adapt to the dynamic nature of real-world groups. 
Even when cost-efficient clustering techniques are adopted \cite{yalcin2021entropy,chen2018hybrid,nozari2020novel,seo2021group}, they often fail to capture the high-order connectivity essential for accurate recommendations. 
Deep learning-based methods can automate group formation during training, and improve recommendation accuracy \cite{kumar2022ophaencoder,liu2024identify}, but they often ignore the underlying graph structure of user interactions and require retraining whenever conditions change. 
Meeting all three requirements simultaneously therefore remains a non-trivial challenge.

To address this gap, we propose \textsf{\textbf{DeepForm}} (\textsf{Stochastic \textbf{Deep} Graph Clustering for Practical Group \textbf{Form}ation}), a deep graph clustering framework specifically designed for GRSs.
\textsf{\textbf{DeepForm}} integrates a lightweight Graph Convolutional Network (GCN)-based deep graph clustering architecture to effectively capture high-order user graph structures (Section \ref{subsec:gae}), and a stochastic cluster learning mechanism to enable dynamic group reconfiguration at inference time without retraining (Section \ref{subsec:scl}).
In addition, contrastive learning is employed to enhance intra-group coherence and inter-group distinction, supporting a wide range of group sizes (Section \ref{subsec:ccl}).
In this way, \textsf{\textbf{DeepForm}} satisfies all three key requirements, ensuring practical group formation for GRSs and providing recommendations that satisfy most group members. 

Our contributions are as follows:
\begin{itemize}[leftmargin=*, topsep=0pt]
    \item \textbf{Discovery.} We identify three key challenges for effective group formation in practical GRSs: 
    (1) incorporation of high-order graph structures, 
    (2) real-time group formation, and
    (3) determination of the optimal number of groups. 
    \item \textbf{Method.} We propose \textsf{\textbf{DeepForm}}, a deep graph clustering framework that jointly addresses all of these challenges.
    \item \textbf{Proof.} Extensive experiments show that \textsf{\textbf{DeepForm}} consistently outperforms existing methods in group formation quality, efficiency, and recommendation accuracy across diverse settings.
\end{itemize}

\begin{table*}[t]
\centering
\setlength{\tabcolsep}{5pt}
\caption{Comparison of related studies, including \textsf{\textbf{DeepForm}}, against the three requirements.}
\vspace{-0.5em}
\label{tab:limit_compare} 
\resizebox{1\textwidth}{!}{
\begin{tabular}{ccccccc}
\toprule
\multicolumn{2}{c}{\textbf{Category}} & \textbf{Model} & \textbf{Group formation method} & \textbf{Graph usage} & \textbf{Real-time} & \textbf{Adjustable $K$} \\
\cmidrule(r{0.5mm}){1-2} \cmidrule(l{0.5mm}r{0.5mm}){3-4} \cmidrule(l{0.5mm}){5-7}
\multirow{12}{*}{GRS}
& \multirow{3}{*}{Predefined}
    &   CubeRec \cite{chen2022thinking}  & Dataset labels & None & Static & Static \\
    & & GroupIM \cite{sankar2020groupim} & Dataset labels & None & Static & Static \\
    & & ConsRec \cite{wu2023consrec}     & Dataset labels & None & Static & Static \\ 
\cmidrule(l{0.5mm}r{0.5mm}){2-2} \cmidrule(l{0.5mm}r{0.5mm}){3-4} \cmidrule(l{0.5mm}r{0.5mm}){5-7}
& \multirow{2}{*}{Heuristic}
    &   \cite{ceh2022performance}     & Cosine similarity & None & Offline & Offline \\
    & & \cite{stratigi2022sequential} & Pairwise scores   & None & Offline & Offline \\ 
\cmidrule(l{0.5mm}r{0.5mm}){2-2} \cmidrule(l{0.5mm}r{0.5mm}){3-4} \cmidrule(l{0.5mm}r{0.5mm}){5-7}
& \multirow{4}{*}{Clustering}
    &   \cite{chen2018hybrid}       & Probabilistic (GMM + social MF)  &  None & Offline & Offline \\
    & & \cite{nozari2020novel}      & Fuzzy C-Means                    &  None  & Offline & Offline \\
    & & \cite{seo2021group}         & Centroid ($K$-Means + genres)      &  None  & Offline & Offline \\
    & & \cite{yalcin2021entropy}    & Info-theoretic (entropy $K$-Means) &  None  & Offline & Offline \\
\cmidrule(l{0.5mm}r{0.5mm}){2-2} \cmidrule(l{0.5mm}r{0.5mm}){3-4} \cmidrule(l{0.5mm}r{0.5mm}){5-7}
& \multirow{2}{*}{\makecell{Deep embedding-based}}
    &   OPHAEncoder \cite{kumar2022ophaencoder} & Deep-latent (AE + $K$-Means)       &  None  & Retrain & Retrain \\
    & & ITR \cite{liu2024identify}      & Density-adaptive (merge/split)   &  None  & Retrain & Retrain \\
\cmidrule(r{0.5mm}){1-2} \cmidrule(l{0.5mm}r{0.5mm}){3-4} \cmidrule(l{0.5mm}){5-7}
\multirow{3}{*}{Non-GRS}
& \multirow{3}{*}{\makecell{Deep graph clustering}}
    &   SDCN \cite{bo2020structural} & Attribute + structure learning & Multi-hop & Offline & Static \\
    & & DAGC \cite{peng2022deep}     & Dual self-supervision          & Multi-hop & Offline & Static \\ 
    & & MGCN \cite{li2024multi}      & Multi-scale learning           & Multi-hop & Offline & Static \\
\cmidrule(r{0.5mm}){1-2} \cmidrule(l{0.5mm}r{0.5mm}){3-4} \cmidrule(l{0.5mm}){5-7}
\first{\textbf{GRS}} & \first{\textbf{Deep graph clustering}} & \first{\textsf{\textbf{DeepForm}}} & \first{\textbf{Stochastic cluster learning}} & \first{\textbf{Multi-hop}} & \first{\textbf{Inference}} & \first{\textbf{Inference}} \\ 
\bottomrule
\end{tabular}
}
\vspace{1em}
\end{table*}

\section{Background}
\label{sec:02_background}

This section explains the process of GRSs and the practical requirements of group formation, laying the groundwork for \textsf{\textbf{DeepForm}}.

\subsection{Group Recommender Systems}
\label{subsec:grs_process}

GRSs operate in two cooperative stages derived from user–item interaction data: \textbf{\textit{\textsf{(1) Group formation}}}, which determines user membership, and \textbf{\textit{\textsf{(2) Group recommendation}}}, which selects items for each group \cite{ceh2022performance,stratigi2022sequential}. 
We formalize each stage and explain \textbf{\textit{\textsf{(3) Why group formation matters}}} for practical deployment, even though it is often treated as secondary in GRSs.

\subsubsection*{\textbf{(1) Group formation.}}
Let $\mathcal{U}$ and $\mathcal{I}$ denote the sets of users and items, respectively.
For effective group formation in practical settings, the process must allow the number of groups $K$ to adapt to dynamic real-world conditions. 
After determining the optimal $K$, the user set $\mathcal{U}$ is partitioned into $K$ non-overlapping groups $\Phi(K) = \mathcal{G}$, where $\Phi$ is the group formation function and $|\mathcal{G}|=K$. 
In real-world settings, group formation is inherently dynamic and context-sensitive, as constraints, resources, and user availability change over time.

\subsubsection*{\textbf{(2) Group recommendation.}}
In the second stage, the GRS either aggregates the individual preferences of group members or learns explicit group representations. 
Aggregation can rely on simple, statistic-based strategies (e.g., Average, Borda Count, Least Misery) \cite{ceh2022performance,baltrunas2010group,yannam2023enhancing,pujahari2015group} or on learned fusion modules \cite{cao2018attentive,cao2019social,ceh2022performance}.
A recommendation model $R$ then produces the recommendation list $R(g) \subseteq \mathcal{I}$ for each group $g \in \mathcal{G}$.
Unlike personalized recommendation, which optimizes for individual satisfaction, group recommendation aims to balance diverse member preferences, prioritizing majority agreement to maximize overall group satisfaction.

\subsubsection*{\textbf{(3) Why group formation matters.}}
Most prior work treats group formation as a disposable pre‑processing stage, focusing primarily on the group recommendation performance.
In practice, however, it is a mission‑critical decision stage. 
While grouping similar users improves recommendation accuracy \cite{krouska2023novel}, this must be balanced with the need for group compositions to adapt to constantly changing real-world conditions \cite{liu2024identify}.
Consequently, enabling dynamic group formation in response to these shifts is crucial for enhancing overall recommendation performance.
Without a fast, standalone group formation module, even state‑of‑the‑art GRSs risk becoming operationally irrelevant in dynamic environments.

\subsection{Requirements of Group Formation}
\label{subsec:requirements}

We identify three interrelated requirements for practical group formation in production GRSs:
\textbf{\textit{\textsf{(1) Incorporation of high-order graph structures}}} to leverage multi-hop connectivity in user–item interactions,
\textbf{\textit{\textsf{(2) Real-time group formation}}} to respond to rapidly changing contexts and constraints, and
\textbf{\textit{\textsf{(3) Dynamic determination of the optimal $K$}}} without retraining or re-optimization.
These requirements stem from operational constraints and directly inform the modeling choices for \textsf{\textbf{DeepForm}}.

\subsubsection*{\textbf{(1) Incorporation of high-order graph structures.}}
Treating users as independent feature vectors discards valuable multi-hop relationships in the user–item interaction graph \cite{wu2022graph}. 
Effective group formation must exploit high-order connectivity while remaining robust to sparsity. 
As modern recommender systems (RSs) increasingly rely on graph-based signals, integrating structural information is foundational for accurate recommendations \cite{anand2025survey}.

\subsubsection*{\textbf{(2) Real-time group formation.}}
Operational constraints, such as maximum group size and the number of groups, can change rapidly. 
A practical GRS should be able to reconfigure both number of groups and memberships at inference time without retraining or service downtime.
Methods that rely on fixed groups or training-time clustering freeze group assignments, forcing costly re-optimization when conditions change \cite{liu2024identify}.
To be viable in real-world deployments, group formation must enable lightweight, inference-time updates under shifting constraints.

\subsubsection*{\textbf{(3) Dynamic determination of the optimal $K$.}}  
The optimal number of groups varies with context, resource limitations, and user composition. 
Therefore, group formation should treat the number of groups as an inference time variable, dynamically adjusting group granularity without retraining. 
This adaptability helps keep group recommendations effective across diverse scenarios.

\section{Related Work}
\label{sec:02_related}

This section analyzes existing group formation methods and deep graph clustering models.
Table \ref{tab:limit_compare} summarizes prior studies in terms of the three requirements for group formation. 
\textbf{\textit{\textsf{Graph usage}}} indicates whether high-order information is incorporated (either \emph{None} or \emph{Multi-hop}). 
\textbf{\textit{\textsf{Real-time}}} refers to how memberships are updated, and \textbf{\textit{\textsf{Adjustable $K$}}} denotes whether the number of groups can change.
Possible states for both are: 
\emph{Static} (fixed from dataset or after training), 
\emph{Retrain} (requiring retraining), \emph{Offline} (updated by re-running on the dataset), and 
\emph{Inference} (adjustable at inference).

\subsection{GRSs Based on Predefined Groups}
\label{subsec:static}

GRSs based on predefined groups treat group membership as fixed labels from the dataset, bypassing the formation process \cite{sankar2020groupim, chen2022thinking, wu2023consrec}.
GroupIM \cite{sankar2020groupim} maximizes mutual information between member and group embeddings to address data sparsity. 
CubeRec \cite{chen2022thinking} represents groups as hypercubes to capture intra-group preference diversity, while ConsRec \cite{wu2023consrec} integrates member-, item-, and group-level signals using hypergraph aggregation.
Although effective on static benchmarks, these methods skip the group formation stage altogether. 
As a result, group compositions remain static, limiting their ability to support adaptive, real-time group reconfigurations. 
Under dynamic real-world conditions, their performance may degrade, requiring costly retraining or fine-tuning \cite{liu2024identify}.

\noindent\textsf{\textit{\textbf{Our Contribution.}}}
We introduce a standalone group formation framework, trained separately and deployed at inference time.
This enables real-time group construction and supports dynamic, practical scenarios without retraining for various group compositions.

\subsection{Heuristic-based Group Formation}
\label{subsec:heuristic}

Some GRSs rely on heuristic-based methods to form groups, 
most often using similarity measures \cite{ceh2022performance,stratigi2022sequential}. 
Ceh et al. \cite{ceh2022performance} instantiate groups by seeding a user and applying cosine-based rules over cluster memberships. 
Stratigi et al. \cite{stratigi2022sequential} implement a formation step in which groups are built by recomputing pairwise similarities and assembling members with high mutual alignment. 
These methods are intuitive and interpretable but have several drawbacks.
They treat users as independent vectors and cannot capture high-order relational structures.
They also require offline similarity recomputation whenever conditions change, making them inefficient for real-time or large-scale deployment.
Finally, they are sensitive to sparsity and noise, often producing unstable or overly homogeneous groups.

\noindent\textsf{\textit{\textbf{Our Contribution.}}}
\textsf{\textbf{DeepForm}} groups users using graph-aware representations that capture high-order structures,
yielding coherent groups while avoiding expensive recomputation at inference.

\subsection{Clustering‑based Group Formation}
\label{subsec:clustering}

To improve scalability, many studies replace pairwise similarity with unsupervised clustering such as $K$‑Means, Gaussian Mixture Models (GMM), or entropy‑aware variants \cite{yalcin2021entropy,chen2018hybrid,nozari2020novel,seo2021group}. 
Chen et al. \cite{chen2018hybrid} fit a GMM over user preference vectors and use the mixture components as groups. 
Nozari et al. \cite{nozari2020novel} apply Fuzzy C-Means on rating similarity to allow partial memberships. 
Yalcin et al. \cite{yalcin2021entropy} employ an entropy-regularized $K$-Means on user–item rating profiles, while Seo et al. \cite{seo2021group} cluster genre-weighted user representations using $K$-Means.
These methods are easy to implement but cannot effectively leverage high-order structural semantics.
As a result, they may produce collapsed clusters and are sensitive to sparsity.
Moreover, applying these methods to the full user-item ratings matrix is inefficient for large-scale deployments.

\noindent\textsf{\textit{\textbf{Our Contribution.}}}
We address the lack of structural awareness by incorporating high-order information through a propagation-only GCN.
In addition, our stochastic cluster learning and contrastive objective make the number of groups adjustable at inference without retraining, preserving the efficiency of clustering methods.

\subsection{Deep Embedding‑based Group Formation} 
\label{subsec:deep_group}

Recent works have explored deep, unsupervised approaches for group formation. OPHAEncoder \cite{kumar2022ophaencoder} combines one-permutation hashing with features from an autoencoder to identify similar users, then merges the results from both to form groups.
ITR \cite{liu2024identify} maps users into a latent space, detects dense regions as centers, and adaptively merges or splits groups during training to produce a final group structure that remains fixed afterward.
However, these approaches cannot adjust the number of groups or update clusters as user behavior changes without full retraining, which limits their practicality in real-world deployments.

\noindent\textsf{\textit{\textbf{Our Contribution.}}}
We extend reconstruction-based training by introducing a joint clustering and contrastive objective that directly enhances cluster quality while leveraging high-order graph structure. 
We also propose stochastic cluster learning, enabling adaptive control over the number of groups at inference without retraining.

\begin{figure*}[t]
    \centering
    \includegraphics[width=0.99\linewidth]{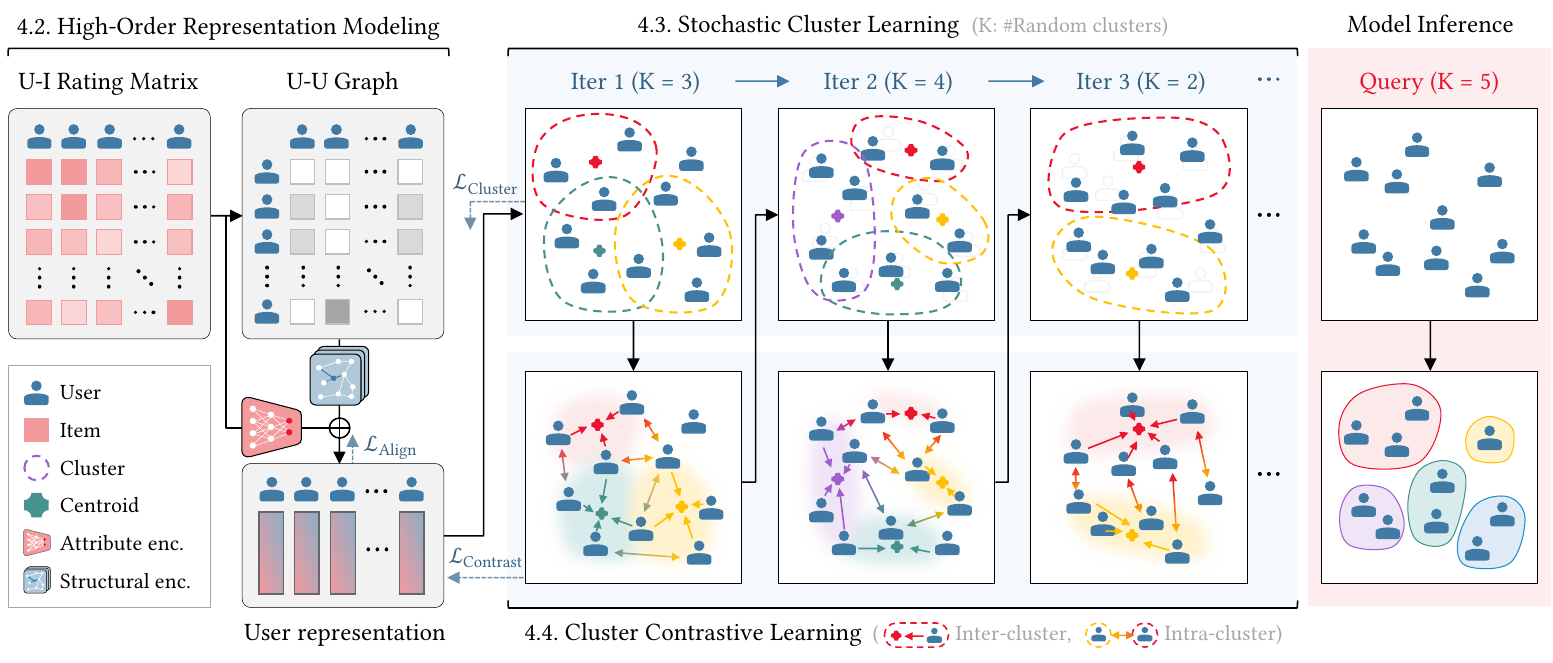}
    \vspace{-0.5em}
    \caption{\textsf{DeepForm} Overview. High-order user representations are obtained from the rating matrix and user graph. Representations are refined in \capblue{\textnormal{\textsf{training}}} via stochastic clustering and contrastive learning and clustered for any desired $K$ at \capred{\textnormal{\textsf{inference}}}.}
    \vspace{0.5em}
\label{fig:model}
\end{figure*}

\subsection{Deep Graph Clustering}
\label{subsec:dgc}

Deep graph clustering networks jointly learn node embeddings and graph partitions.
SDCN \cite{bo2020structural} integrates graph convolutions with an autoencoder to balance reconstruction and intra-cluster coherence.
DAGC \cite{peng2022deep} enhances robustness through edge-level attention and dual self-supervision, while MGCN \cite{li2024multi} extends structural reach via hierarchical pooling across multiple scales. 
However, these methods are designed for static graphs with fixed node sets and assume a predetermined number of groups, which limits their flexibility in dynamic environments \cite{liu2022survey}.
They are not tailored to the demands of practical GRSs, where user–item graphs are sparse, frequently updated, and their group composition can vary for each request.
Applying these models directly often results in unstable clustering, poor generalization, and inefficiency.

\noindent\textsf{\textit{\textbf{Our Contribution.}}}
We present the first deep graph clustering model explicitly designed for GRSs.
Unlike prior work, we adopt a propagation-only architecture, which better preserves high-order structure without introducing unnecessary complexity or overfitting in sparse interaction data.
We also introduce stochastic cluster learning, enabling real-time group formation with a dynamic number of groups at inference.

\section{Methodology}
\label{sec:04_proposed}

This section introduces our method, \textsf{\textbf{DeepForm}} (\textsf{Stochastic \textbf{Deep} Graph Clustering for Practical Group \textbf{Form}ation}).

\subsection{Overview}
\label{subsec:overview}

\textsf{\textbf{DeepForm}} consists of three main modules \textbf{\textit{\textsf{(1) High-order Representation Modeling}}}, \textbf{\textit{\textsf{(2) Stochastic Cluster Learning}}}, and \textbf{\textit{\textsf{(3) Cluster Contrastive Learning}}}, as illustrated in Figure \ref{fig:model}. 
First, a lightweight, propagation-only GCN encodes high-order structural information from user–item interactions (Section \ref{subsec:gae}).
Next, the stochastic cluster learning mechanism enables real-time group formation at inference time (Section \ref{subsec:scl}).
Finally, cluster contrastive learning promotes representational coherence and adaptability under dynamic conditions (Section \ref{subsec:ccl}).
The overall training and inference process of \textsf{\textbf{DeepForm}} is described in Section \ref{subsec:loss}.

\subsection{High-order Representation Modeling}
\label{subsec:gae}

RSs increasingly exploit graph structure to overcome data sparsity and capture implicit relationships \cite{anand2025survey}. In this context, deep graph clustering is well suited to group formation because it integrates graph information directly into the clustering process. However, applying such methods to GRSs requires tailored design and optimization to address their unique challenges. 

Conventional deep graph clustering typically assumes relationships among single-type nodes, but GRSs involve multiple node types, such as users and items, which necessitates an additional approach to integrate them. Nevertheless, we model implicit relations using only user nodes for three reasons. First, item nodes are usually far more numerous than user nodes, and modeling them in detail incurs substantial computational costs. Second, group formation is inherently defined over relationships among users, so user relations alone suffice to capture high-order patterns. Third, focusing the GCN solely on user-user relations enables more effective exploration of latent correlations. Accordingly, we adopt a LightGCN-based architecture \cite{he2020lightgcn}, which is effective for RSs by eliminating nonlinear transformations. 

Formally, we construct the user–user graph $\mathbf{A}$ from the normalized user–item rating matrix 
$\mathbf{X}\in\mathbb{R}^{|\mathcal{U}|\times|\mathcal{I}|}$. 
Let $\mathbf{x}_u \in \mathbb{R}^{|\mathcal{I}|}$ denote user $u$'s row in $\mathbf{X}$. 
Then each entry $a_{uv}$ of $\mathbf{A}$ is calculated by dot-product similarity between users $u$ and $v$, defined as:
\begin{equation}
    a_{uv} = 
    \begin{cases}
        \max(\mathbf{x}_u^\top \mathbf{x}_v, 0), & u \neq v, \\
        0, & u = v.
    \end{cases}
\end{equation}
Using the $d$-dimensional user embedding $\mathbf{Z}\in\mathbb{R}^{|\mathcal{U}|\times d}$, 
the forward and backward GCN representations after aggregating $L$ hops are:
\begin{equation}
    \mathbf{Z}_{\text{GCN}} = \frac{1}{L+1} \sum_{l=0}^{L} \tilde{\mathbf{A}}^{l}\mathbf{Z}, 
    \quad
    \mathbf{\hat{Z}} = \frac{1}{L+1} \sum_{l=0}^{L} \tilde{\mathbf{A}}^{l}\mathbf{Z}_{\text{GCN}},
\label{eq:gcn-combined}
\end{equation}
where $\tilde{\mathbf{A}}=\mathbf{D}^{-1/2}(\mathbf{A}+\mathbf{I})\mathbf{D}^{-1/2}$ is the symmetrically normalized user–user adjacency matrix.
We use the backward GCN representation $\mathbf{\hat{Z}}$ for reconstruction.

While LightGCN can effectively capture structural user-user relationships, it may still lack complementary information arising from user–item interactions. To address this, we incorporate an autoencoder (AE)-based representation using the user-item rating matrix. In existing deep graph clustering, AEs are used to model node features directly. However, in GRSs, items do not have independent features beyond indicating user preferences. Since user relationships have already been modeled through the graph structure, we transform these interactions into a user-item rating matrix and compress it with an AE to supplement information. This enables \textsf{\textbf{DeepForm}} to integrate item-related context into user-level structural representations, thereby producing high-order graph representations that are effective for group formation.

We employ a multi-layer perceptron (MLP) with three layers $\mathcal{M}$ as the encoder and $\mathcal{M}^{-1}$ as the decoder:
\begin{align}
    \mathbf{Z}_{\text{AE}} = \mathcal{M}(\mathbf{X}), \quad
    \hat{\mathbf{X}}_{\text{AE}} = \mathcal{M}^{-1}(\mathbf{Z}_{\text{AE}}),
\label{eq:ae}
\end{align}

In summary, our deep graph clustering method for GRSs achieves effective modeling with relatively low computational cost. During training, unlike traditional modular approaches \cite{bo2020structural, peng2022deep}, we integrate the AE as an auxiliary learning term within the GCN training process. This joint approach preserves consistency between the structural and attribute representations, ensures that clustering remains focused on user relationships, and stably improves the quality of group formation. 

The training objective is defined as:
\begin{equation}
\mathcal{L}_{\text{Align}}
    \!=\!\underbrace{\lVert\hat{\mathbf{Z}}\!-\!\mathbf{Z}\rVert_F^{2}
    +\!\lVert\hat{\mathbf{A}}\!-\!\mathbf{A}\rVert_F^{2}}_{\text{GCN Reconst.}}
    +\underbrace{\lVert\hat{\mathbf{X}}_{\text{AE}}\!-\!\mathbf{X}\rVert_F^{2}}_{\text{AE Reconst.}}
    +\underbrace{\lVert\mathbf{Z}_{\text{GCN}}\!-\!\mathbf{Z}_{\text{AE}}\rVert_F^{2}}_{\text{Alignment}},
\end{equation}
where $\hat{\mathbf{A}} = \sigma\bigl(\mathbf{Z}_{\text{GCN}}\mathbf{Z}^\top_{\text{GCN}}\bigr)$ and $\lVert\cdot\rVert_F$ denote the Frobenius norm.
Scalar weights for balancing each term are omitted for simplicity.

\subsection{Stochastic Cluster Learning}
\label{subsec:scl}

To construct practical applications for GRSs, it is essential to operate in real-time while considering constraints such as resource limitations. However, our deep graph clustering model introduced in Section \ref{subsec:gae}, while computationally efficient, inherently suffers from the limitation that any change in the number of clusters requires retraining. To address this, we introduce a stochastic cluster learning strategy that enables flexible group formation during inference without additional training, while also mitigating overfitting to a specific number of clusters.

Specifically, at each training iteration, the number of clusters $K$ is uniformly sampled from a predefined search space, allowing the model to be exposed to diverse interaction patterns that may arise from various group types. Compared to conventional methods that use a fixed $K$, this stochastic sampling $K \sim \text{Uniform}(2, K_{\max})$ offers the following advantages:
\begin{itemize}[leftmargin=*, topsep=0pt]
    \item \textbf{Real-time capability}: Delivers rapid group formation results without additional retraining, unlike conventional deep graph clustering methods.
    \item \textbf{Constraint adaptability}: Supports dynamic group formation under resource limitations (e.g., maximum group size, number of groups).
    \item \textbf{Performance generalization}: Trains with a variety of group sizes, enabling unbiased inference for unseen numbers of groups.
\end{itemize}
This work integrates the stochastic cluster learning strategy into the existing optimization process, allowing deep graph clustering to support practical group formation in real-world scenarios.

Following \cite{bo2020structural,peng2022deep,li2024multi}, we adopt soft clustering assignments during training based on the final user representation \(\mathbf{Z}=\mathbf{Z}_{\text{GCN}}+\mathbf{Z}_{\text{AE}}\).
For each user embedding $\mathbf{z}_u \in \mathbb{R}^d$, 
we compute its similarity score $q_{uc}$ by normalizing the Student’s $t$-distribution kernel 
$\kappa(u,c) = \left(1 + \lVert \mathbf{z}_u - \boldsymbol{\mu}_c \rVert_2^2\right)^{-1}$ 
between each user embedding $\mathbf{z}_u$ and cluster centroid 
$\{\boldsymbol{\mu}_1, \dots, \boldsymbol{\mu}_K\}$, 
and construct the target distribution $p_{uc}$ by sharpening confident assignments:
\begin{equation}
q_{uc} = \operatorname{Norm}\bigl[\kappa(u,c)\bigr],
\quad
p_{uc} = \operatorname{Norm}\left[\frac{q_{uc}^2}{\sum_{i=1}^{|\mathcal{U}|} q_{ic}}\right],
\end{equation}
where $\operatorname{Norm}[x_c]=\frac{x_c}{\sum_{j=1}^K x_j}$ is a cluster-wise normalization operator.

The final cluster loss, which aligns the soft cluster assignments $q$ with the sharpened target distribution $p$, is defined using the Kullback–Leibler divergence (KLD) as follows:
\begin{equation}
\mathcal{L}_{\mathrm{Cluster}}
= \sum_{u=1}^{|\mathcal{U}|}\sum_{c=1}^{K} p_{uc}\log\!\frac{p_{uc}}{q_{uc}}.
\end{equation}

This ensures the formation of more distinct and well-separated clusters for each of the group numbers sampled during training, enhancing the model’s generalization across different group sizes.

\subsection{Cluster Contrastive Learning}
\label{subsec:ccl}

The components introduced above support the intrinsic quality and generalization of user representations across various $K$ values, but these representations can be unstable because cluster boundaries continuously change during training \cite{leiber2024dying}. 
To ensure both intra-cluster cohesion and inter-cluster separation under such variability, we design a cluster contrastive learning objective composed of two complementary losses: a triplet loss and an InfoNCE loss.
Whenever a new $K$ is sampled, we update clusters by running $K$-Means on the current user embeddings before computing the contrastive losses.

First, a triplet loss sharpens inter-cluster separation by anchoring users near their assigned cluster centers at each training stage \cite{dong2018triplet}.
For an anchor user $u$ assigned to cluster $c$, a positive center $\boldsymbol{\mu}_c$, and a negative user $u^- \in \mathcal{U}$ from a different cluster, we define:
\begin{equation}
    \mathcal{L}_{\text{Triplet}}
    = \max\!\bigl\{0,\,
        \lVert\mathbf{z}_u-\boldsymbol{\mu}_{c}\rVert_2
      - \lVert\mathbf{z}_{u^-}-\boldsymbol{\mu}_{c}\rVert_2
      + \Delta\bigr\}.     
\label{eq:triplet}
\end{equation}
This loss pulls members closer to their own center while enforcing a minimum distance $\Delta$ from users outside the cluster.

Second, an InfoNCE loss promotes intra-cluster cohesion \cite{yannam2023enhancing}.
For an anchor user $u$, a positive user $u^+$ from the same provisional cluster, and negatives $u^- \in \mathcal{N}(u)$, we define $\mathcal{L}_{\text{NCE}}$:
\begin{equation}
    \mathcal{L}_{\text{NCE}}
      = -\log \frac{\exp\!\bigl((\mathbf{z}_u^{\!\top}\mathbf{z}_{u^+})/\tau\bigr)}{\sum_{u^-}\exp\!\bigl((\mathbf{z}_u^{\!\top}\mathbf{z}_{u^-})/\tau\bigr)},
\label{eq:nce}
\end{equation}
where $\tau$ is a temperature parameter that controls the concentration of similarity.
Minimizing this loss increases intra-cluster cohesion while keeping clusters separated.

Simultaneous optimization of $\mathcal{L}_{\text{Contrast}}=\mathcal{L}_{\text{Triplet}}+\mathcal{L}_{\text{NCE}}$ yields embeddings that are both compact within clusters and well-separated across clusters, promoting stable group assignments under dynamic conditions.  
Moreover, the hierarchical structure induced by stochastic cluster learning allows the model to preserve consistent relational geometry across multiple granularities, enabling the resulting embeddings to support a broad range of group counts $K$ without retraining.

\subsection{Model Training and Inference}
\label{subsec:loss}

The final loss function integrates all loss components to ensure robust clustering and effective user representation learning. All terms combine into:
\begin{equation}
    \mathcal{L}_{\text{Final}}
      = \mathcal{L}_{\text{Align}}
      + \mathcal{L}_{\text{Cluster}}
      + \mathcal{L}_{\text{Contrast}}.
\label{eq:total}
\end{equation}
Optimizing the final loss $\mathcal{L}_{\text{Final}}$ produces user embeddings that 
capture high-order structure from the user–item interactions ($\mathcal{L}_{\text{Align}}$), 
enable efficient, real-time group formation at inference ($\mathcal{L}_{\text{Cluster}}$), and 
support dynamic adjustment of the number of groups without retraining ($\mathcal{L}_{\text{Contrast}}$).
These properties directly satisfy the core requirements for practical group formation in GRSs.

During inference, we use the final user embeddings $\mathbf{Z}$ optimized by minimizing $\mathcal{L}_{\text{Final}}$ and apply a $K$-Means group formation function to form the final groups $\mathcal{G}_\text{opt}$ for any given $K$:
\begin{equation}
    \mathcal{G}_\text{opt} = \arg\min_{\mathcal{G}} \sum_{g \in \mathcal{G}} \sum_{u \in g} \|\mathbf{z}_u - \boldsymbol{\mu}_g\|_2^2,
\end{equation}
where $\boldsymbol{\mu}_g$ is the centroid of group $g \in \mathcal{G}$. 
This formulation enables dynamic group reconfiguration during inference for any desired $K$, eliminating the need for retraining while incorporating the high-order structure and contrastive separation learned during training.

\section{Experimental Setup}
\label{sec:05_exp_setup}

This section outlines the experimental setup used for assessing \textsf{\textbf{DeepForm}}. 
It describes the datasets, evaluation metrics, baselines, and implementation details used in the study.

\begin{table}[t]
\renewcommand{\arraystretch}{1}
\caption{Statistics of datasets used in our experiments.}
\vspace{-0.5em}
\label{tab:datasets}
\setlength{\tabcolsep}{3.5pt}
\centering
\resizebox{1\columnwidth}{!}{
\begin{tabular}{lccccc} 
    \toprule
    & \multicolumn{3}{c}{\textbf{Entities}} & \multicolumn{2}{c}{\textbf{Interactions}} \\
    \cmidrule(l{0.5mm}r{0.5mm}){2-4} \cmidrule(l{0.5mm}){5-6}
    \textbf{Dataset} & \textbf{Users} & \textbf{Items} & \textbf{Groups} & \textbf{User-Item} & \textbf{Group-Item} \\
    \cmidrule(r{0.5mm}){1-1} \cmidrule(l{0.5mm}r{0.5mm}){2-4}  \cmidrule(l{0.5mm}){5-6}
    Baby        & 19,445 & 7,050  & --  & 160,792 & -- \\
    Clothing    & 39,387 & 23,033 & --  & 278,677 & -- \\
    \cmidrule(r{0.5mm}){1-1} \cmidrule(l{0.5mm}r{0.5mm}){2-4}  \cmidrule(l{0.5mm}){5-6}
    CAMRa2011   &   602  & 7,710  & 290 & 116,344 & 145,068 \\
    \bottomrule
\end{tabular}
}
\end{table}

\subsubsection*{\textbf{Datasets.}} 
We evaluate our method on three widely used real-world datasets that cover both ad-hoc and predefined group scenarios, enabling comprehensive comparisons across diverse group recommendation environments. 
Dataset statistics are summarized in Table \ref{tab:datasets}.
To evaluate general performance in group formation scenarios, we use the Amazon Baby and Clothing\footnote{\url{https://jmcauley.ucsd.edu/data/amazon/}}, which contain user–item interactions.
These settings emulate real-world scenarios where groups must be inferred from individual data.
Both datasets are standard benchmarks in GRSs, and are preprocessed to retain users and items with at least 10 interactions for meaningful comparisons.
Furthermore, to compare performance between predefined and ad-hoc formed groups, we employ CAMRa2011\footnote{\url{https://recsys.acm.org/recsys11/camra/}}, which includes explicit group labels and group-level interactions. 
This allows analysis of how variations in group formation affect overall effectiveness.

\subsubsection*{\textbf{Evaluation Metrics.}} 
We evaluate model performance using two standard metrics widely adopted in GRSs: Normalized Discounted Cumulative Gain (NDCG) \cite{xu2024aligngroup,sankar2020groupim,chen2022thinking,wu2023consrec,ceh2022performance,stratigi2022sequential,kumar2022ophaencoder,liu2024identify,yalcin2021entropy,seo2021group} and Hit Rate (HR) \cite{xu2024aligngroup,wu2023consrec,ceh2022performance,liu2024identify}. 
Together, these metrics capture both the relevance and ranking quality of recommended items. 
Results are reported for top-$k$ with $k \in \{5, 10, 20\}$ to assess effectiveness across different numbers of recommended items.

\subsubsection*{\textbf{Baselines.}}
To evaluate the effectiveness of \textsf{\textbf{DeepForm}}, we compare its performance across three categories of baselines under different desired group sizes:
heuristic-based (\textbf{Similarity}), classical clustering-based (\textbf{$K$-Means} and \textbf{GMM}), and deep graph clustering-based methods (\textbf{SDCN}, \textbf{DAGC}, and \textbf{MGCN}).
While deep embedding-based methods were potential candidates, their fixed group size—regardless of desired size—makes them unsuitable for fair comparison.
Detailed descriptions of each baseline are as follows:

\begin{itemize}[leftmargin=*]
    \item \textbf{Similarity} \cite{ceh2022performance}: Users are grouped based on pairwise similarity derived from the ratings matrix using Pearson correlation.
    \item \textbf{$K$-Means} \cite{seo2021group,yalcin2021entropy}: A widely used clustering algorithm that clusters users based on feature similarity in Euclidean space.
    \item \textbf{GMM} \cite{chen2018hybrid}: A probabilistic model that represents the data as a mixture of Gaussian distributions.
    \item \textbf{SDCN} \cite{bo2020structural}: A deep graph clustering model that employs two parallel subnetworks, jointly leveraging node attributes and structural information.
    \item \textbf{DAGC} \cite{peng2022deep}: Incorporates a dual self-supervision strategy to refine cluster structures and enhances model robustness through feature fusion using attention mechanisms.
    \item \textbf{MGCN} \cite{li2024multi}: A multi-scale graph clustering framework that integrates structural and attribute information across multiple granularities in the latent space. 
\end{itemize}

\subsubsection*{\textbf{Group Recommender Algorithms.}}
To assess the impact of group formation on recommendation performance, we employ both representative aggregation strategies and neural approaches. The aggregation strategies include three widely used methods: Average (\textbf{AVG}), Borda Count (\textbf{BC}), and Least Misery (\textbf{LM}). For the neural approach, we adopt an algorithm based on Neural Collaborative Filtering (\textbf{NCF}).

\begin{itemize}[leftmargin=*]
    \item \textbf{AVG} \cite{ceh2022performance}: Computes the average rating for each item across all group members, and ranks items by the average.
    \item \textbf{BC} \cite{baltrunas2010group,ceh2022performance,zhu2020context}: Assigns item scores based on their rank in each user’s preference list. 
    The scores are cumulated into total scores. 
    \item \textbf{LM} \cite{ceh2022performance,pujahari2015group,ghazarian2015enhancing}: Ranks items based on the lowest individual rating within the group, emphasizing fairness by avoiding items strongly disliked by any member.
    \item \textbf{NCF} \cite{ceh2022performance,cao2018attentive,cao2019social}: Constructs a group embedding by averaging NCF-based user embeddings, enabling neural scoring of items based on learned group-level representations.
\end{itemize}

\subsubsection*{\textbf{Implementation Details.}} 
Experiments are conducted in an Ubuntu 20.04 environment using an AMD EPYC 7513 2.6GHz CPU and an NVIDIA GeForce RTX 4090 GPU with Python 3.11, PyTorch 2.4.1, and CUDA 12.4. 
The learning rate of the proposed model is set to $1 \times 10^{-5}$, and the model is trained for 200 epochs. 
The deep baseline models are trained using the hyperparameters from the authors’ official implementations.

\begin{table*}[t]
\centering
\caption{Performance comparison with group number $K=128$. The best and second-best results are shown in \capfirst{bold} and \underline{underlined}.
N denotes NDCG and H denotes HR. Higher values indicate better performance.}
\vspace{-9pt}

\begin{subtable}{\textwidth}
\renewcommand{\arraystretch}{0.9}
\centering
\subcaption{\textnormal{AVG: \textsf{\textbf{DeepForm}} ranks first in every column, indicating consistent gains.}}
\vspace{-6pt}
\setlength{\tabcolsep}{7pt}

\resizebox{\textwidth}{!}{
\begin{tabular}{lcccccccccccc}
    \toprule
    & \multicolumn{6}{c}{\textbf{Baby}} & \multicolumn{6}{c}{\textbf{Clothing}} \\
    \cmidrule(l{1mm}r{1mm}){2-7} \cmidrule(l{1mm}){8-13}
    \textbf{Method} & \textbf{N@5} & \textbf{N@10} & \textbf{N@20} & \textbf{H@5} & \textbf{H@10} & \textbf{H@20} & \textbf{N@5} & \textbf{N@10} & \textbf{N@20} & \textbf{H@5} & \textbf{H@10} & \textbf{H@20} \\
    \cmidrule(r{1mm}){1-1} \cmidrule(l{1mm}r{1mm}){2-4} \cmidrule(l{1mm}r{1mm}){5-7} \cmidrule(l{1mm}r{1mm}){8-10} \cmidrule(l{1mm}){11-13} 
    Similarity & 0.0082 & 0.0117 & 0.0169 & 0.0267 & 0.0493 & 0.0858 & 0.0052 & 0.0068 & 0.0093 & 0.0121 & 0.0201 & 0.0344 \\
    \cmidrule(r{1mm}){1-1} \cmidrule(l{1mm}r{1mm}){2-4} \cmidrule(l{1mm}r{1mm}){5-7} \cmidrule(l{1mm}r{1mm}){8-10} \cmidrule(l{1mm}){11-13}
    $K$-Means    & 0.0055 & 0.0084 & 0.0139 & 0.0202 & 0.0395 & 0.0791 & 0.0035 & 0.0048 & 0.0070 & 0.0108 & 0.0191 & 0.0318 \\
    GMM        & 0.0082 & 0.0118 & 0.0164 & 0.0263 & 0.0494 & 0.0845 & 0.0050 & 0.0067 & 0.0092 & 0.0117 & 0.0197 & 0.0341 \\ 
    \cmidrule(r{1mm}){1-1} \cmidrule(l{1mm}r{1mm}){2-4} \cmidrule(l{1mm}r{1mm}){5-7} \cmidrule(l{1mm}r{1mm}){8-10} \cmidrule(l{1mm}){11-13}
    SDCN       & 0.0086 & 0.0124 & 0.0179 & 0.0316 & \underline{0.0626} & \underline{0.1099} & 0.0060 & 0.0080 & 0.0105 & 0.0164 & 0.0293 & 0.0436 \\
    DAGC       & 0.0084 & 0.0115 & 0.0162 & 0.0294 & 0.0569 & 0.1006 & 0.0056 & 0.0069 & 0.0099 & 0.0132 & 0.0218 & 0.0370 \\
    MGCN       & \underline{0.0091} & \underline{0.0129} & \first{0.0185} & \underline{0.0324} & 0.0623 & 0.1073 & \underline{0.0067} & \underline{0.0092} & \underline{0.0121} & \underline{0.0170} & \underline{0.0312} & \underline{0.0486} \\ 
    \cmidrule(r{1mm}){1-1} \cmidrule(l{1mm}r{1mm}){2-4} \cmidrule(l{1mm}r{1mm}){5-7} \cmidrule(l{1mm}r{1mm}){8-10} \cmidrule(l{1mm}){11-13}
    \first{\textsf{\textbf{DeepForm}}}       & \first{0.0098} & \first{0.0136} & \first{0.0185} & \first{0.0365} & \first{0.0678} & \first{0.1133} & \first{0.0080} & \first{0.0104} & \first{0.0136} & \first{0.0214} & \first{0.0365} & \first{0.0547} \\
    \bottomrule
\end{tabular}
}
\end{subtable}

\vspace{4pt}

\begin{subtable}{\textwidth}
\renewcommand{\arraystretch}{0.9}
\centering
\subcaption{\textnormal{BC: \textsf{\textbf{DeepForm}} achieves the strongest overall performance, but MGCN achieves the best N@5 and N@20 on Baby.}}
\vspace{-6pt}
\setlength{\tabcolsep}{7pt}
\resizebox{\textwidth}{!}{
\begin{tabular}{lcccccccccccc}
    \toprule
    & \multicolumn{6}{c}{\textbf{Baby}} & \multicolumn{6}{c}{\textbf{Clothing}} \\
    \cmidrule(l{1mm}r{1mm}){2-7} \cmidrule(l{1mm}){8-13}
    \textbf{Method} & \textbf{N@5} & \textbf{N@10} & \textbf{N@20} & \textbf{H@5} & \textbf{H@10} & \textbf{H@20} & \textbf{N@5} & \textbf{N@10} & \textbf{N@20} & \textbf{H@5} & \textbf{H@10} & \textbf{H@20} \\
    \cmidrule(r{1mm}){1-1} \cmidrule(l{1mm}r{1mm}){2-4} \cmidrule(l{1mm}r{1mm}){5-7} \cmidrule(l{1mm}r{1mm}){8-10} \cmidrule(l{1mm}){11-13}
    Similarity & 0.0076 & 0.0115 & 0.0165 & 0.0248 & 0.0476 & 0.0836 & 0.0051 & 0.0067 & 0.0094 & 0.0120 & 0.0193 & 0.0349 \\
    \cmidrule(r{1mm}){1-1} \cmidrule(l{1mm}r{1mm}){2-4} \cmidrule(l{1mm}r{1mm}){5-7} \cmidrule(l{1mm}r{1mm}){8-10} \cmidrule(l{1mm}){11-13}
    $K$-Means    & 0.0046 & 0.0092 & 0.0140 & 0.0168 & 0.0404 & 0.0787 & 0.0030 & 0.0048 & 0.0070 & 0.0076 & 0.0192 & 0.0318 \\
    GMM        & 0.0077 & 0.0114 & 0.0163 & 0.0251 & 0.0476 & 0.0828 & 0.0051 & 0.0066 & 0.0093 & 0.0121 & 0.0194 & 0.0344 \\ 
    \cmidrule(r{1mm}){1-1} \cmidrule(l{1mm}r{1mm}){2-4} \cmidrule(l{1mm}r{1mm}){5-7} \cmidrule(l{1mm}r{1mm}){8-10} \cmidrule(l{1mm}){11-13}
    SDCN       & \underline{0.0082} & 0.0121 & 0.0175 & 0.0290 & \underline{0.0599} & \underline{0.1066} & 0.0059 & 0.0079 & 0.0106 & 0.0167 & 0.0284 & 0.0451 \\
    DAGC       & 0.0076 & 0.0120 & 0.0165 & 0.0244 & 0.0560 & 0.0996 & 0.0056 & 0.0070 & 0.0097 & 0.0138 & 0.0221 & \underline{0.0372} \\
    MGCN       & \first{0.0087} & \underline{0.0127} & \first{0.0184} & \first{0.0311} & 0.0599 & 0.1050 & \underline{0.0068} & \underline{0.0090} & \underline{0.0123} & \underline{0.0173} & \underline{0.0300} & 0.0502 \\ 
    \cmidrule(r{1mm}){1-1} \cmidrule(l{1mm}r{1mm}){2-4} \cmidrule(l{1mm}r{1mm}){5-7} \cmidrule(l{1mm}r{1mm}){8-10} \cmidrule(l{1mm}){11-13}
    \first{\textsf{\textbf{DeepForm}}}       & 0.0081 & \first{0.0134} & \underline{0.0183} & \underline{0.0294} & \first{0.0659} & \first{0.1098} & \first{0.0079} & \first{0.0103} & \first{0.0137} & \first{0.0207} & \first{0.0359} & \first{0.0552} \\
    \bottomrule
\end{tabular}
}
\end{subtable}

\vspace{4pt}

\begin{subtable}{\textwidth}
\renewcommand{\arraystretch}{0.9}
\centering
\subcaption{\textnormal{LM: \textsf{\textbf{DeepForm}} consistently outperforms all baselines across all metrics.}}
\vspace{-6pt}
\setlength{\tabcolsep}{7pt}
\resizebox{1\textwidth}{!}{
\begin{tabular}{lcccccccccccc}
    \toprule
    & \multicolumn{6}{c}{\textbf{Baby}} & \multicolumn{6}{c}{\textbf{Clothing}} \\
    \cmidrule(l{1mm}r{1mm}){2-7} \cmidrule(l{1mm}){8-13}
    \textbf{Method} & \textbf{N@5} & \textbf{N@10} & \textbf{N@20} & \textbf{H@5} & \textbf{H@10} & \textbf{H@20} & \textbf{N@5} & \textbf{N@10} & \textbf{N@20} & \textbf{H@5} & \textbf{H@10} & \textbf{H@20} \\
    \cmidrule(r{1mm}){1-1} \cmidrule(l{1mm}r{1mm}){2-4} \cmidrule(l{1mm}r{1mm}){5-7} \cmidrule(l{1mm}r{1mm}){8-10} \cmidrule(l{1mm}){11-13}
    Similarity & 0.0084 & 0.0114 & 0.0167 & 0.0270 & 0.0489 & 0.0861 & 0.0049 & 0.0067 & 0.0089 & 0.0115 & 0.0201 & 0.0336 \\
    \cmidrule(r{1mm}){1-1} \cmidrule(l{1mm}r{1mm}){2-4} \cmidrule(l{1mm}r{1mm}){5-7} \cmidrule(l{1mm}r{1mm}){8-10} \cmidrule(l{1mm}){11-13}
    $K$-Means    & 0.0055 & 0.0083 & 0.0137 & 0.0206 & 0.0394 & 0.0786 & 0.0034 & 0.0048 & 0.0069 & 0.0108 & 0.0191 & 0.0318 \\
    GMM        & 0.0083 & 0.0116 & 0.0164 & 0.0268 & 0.0494 & 0.0859 & 0.0047 & 0.0064 & 0.0090 & 0.0114 & 0.0198 & 0.0341 \\ 
    \cmidrule(r{1mm}){1-1} \cmidrule(l{1mm}r{1mm}){2-4} \cmidrule(l{1mm}r{1mm}){5-7} \cmidrule(l{1mm}r{1mm}){8-10} \cmidrule(l{1mm}){11-13}
    SDCN       & 0.0088 & 0.0120 & 0.0177 & 0.0328 & 0.0626 & \underline{0.1093} & 0.0061 & 0.0079 & 0.0104 & 0.0166 & 0.0286 & 0.0437 \\
    DAGC       & 0.0083 & 0.0112 & 0.0160 & 0.0288 & 0.0554 & 0.1008 & 0.0055 & 0.0068 & 0.0099 & 0.0138 & 0.0218 & 0.0370 \\
    MGCN       & \underline{0.0092} & \underline{0.0127} & \underline{0.0182} & \underline{0.0329} & \underline{0.0627} & 0.1071 & \underline{0.0063} & \underline{0.0089} & \underline{0.0115} & \underline{0.0167} & \underline{0.0308} & \underline{0.0481} \\ 
    \cmidrule(r{1mm}){1-1} \cmidrule(l{1mm}r{1mm}){2-4} \cmidrule(l{1mm}r{1mm}){5-7} \cmidrule(l{1mm}r{1mm}){8-10} \cmidrule(l{1mm}){11-13}
    \first{\textsf{\textbf{DeepForm}}}        & \first{0.0098} & \first{0.0135} & \first{0.0185} & \first{0.0371} & \first{0.0680} & \first{0.1144} & \first{0.0074} & \first{0.0099} & \first{0.0132} & \first{0.0210} & \first{0.0353} & \first{0.0551} \\
    \bottomrule
\end{tabular}
}
\end{subtable}

\vspace{4pt}
\begin{subtable}{\textwidth}
\renewcommand{\arraystretch}{0.9}
\centering
\subcaption{\textnormal{NCF: \textsf{\textbf{DeepForm}} achieves the strongest overall performance on all metrics except N@20 on Baby.}}
\vspace{-6pt}
\setlength{\tabcolsep}{7pt}
\resizebox{1\textwidth}{!}{
\begin{tabular}{lcccccccccccc}
    \toprule
    & \multicolumn{6}{c}{\textbf{Baby}} & \multicolumn{6}{c}{\textbf{Clothing}} \\
    \cmidrule(l{1mm}r{1mm}){2-7} \cmidrule(l{1mm}){8-13}
    \textbf{Method} & \textbf{N@5} & \textbf{N@10} & \textbf{N@20} & \textbf{H@5} & \textbf{H@10} & \textbf{H@20} & \textbf{N@5} & \textbf{N@10} & \textbf{N@20} & \textbf{H@5} & \textbf{H@10} & \textbf{H@20} \\
    \cmidrule(r{1mm}){1-1} \cmidrule(l{1mm}r{1mm}){2-4} \cmidrule(l{1mm}r{1mm}){5-7} \cmidrule(l{1mm}r{1mm}){8-10} \cmidrule(l{1mm}){11-13}
    Similarity & 0.0079 & 0.0115 & 0.0165 & 0.0261 & 0.0479 & 0.0835 & 0.0052 & 0.0067 & 0.0094 & 0.0123 & 0.0195 & 0.0348 \\
    \cmidrule(r{1mm}){1-1} \cmidrule(l{1mm}r{1mm}){2-4} \cmidrule(l{1mm}r{1mm}){5-7} \cmidrule(l{1mm}r{1mm}){8-10} \cmidrule(l{1mm}){11-13}
    $K$-Means    & 0.0057 & 0.0085 & 0.0139 & 0.0209 & 0.0391 & 0.0788 & 0.0034 & 0.0048 & 0.0070 & 0.0107 & 0.0190 & 0.0318 \\
    GMM        & 0.0080 & 0.0114 & 0.0164 & 0.0264 & 0.0478 & 0.0833 & 0.0051 & 0.0067 & 0.0093 & 0.0122 & 0.0195 & 0.0346 \\ 
    \cmidrule(r{1mm}){1-1} \cmidrule(l{1mm}r{1mm}){2-4} \cmidrule(l{1mm}r{1mm}){5-7} \cmidrule(l{1mm}r{1mm}){8-10} \cmidrule(l{1mm}){11-13}
    SDCN       & 0.0085 & 0.0121 & 0.0176 & 0.0314 & \underline{0.0608} & \underline{0.1073} & 0.0059 & 0.0079 & 0.0106 & 0.0169 & 0.0284 & 0.0450 \\
    DAGC       & 0.0085 & 0.0121 & 0.0165 & 0.0295 & 0.0571 & 0.0997 & 0.0058 & 0.0070 & 0.0100 & 0.0137 & 0.0221 & 0.0374 \\
    MGCN       & \underline{0.0089} & \underline{0.0128} & \first{0.0185} & \underline{0.0326} & 0.0605 & 0.1061 & \underline{0.0070} & \underline{0.0091} & \underline{0.0123} & \underline{0.0179} & \underline{0.0302} & \underline{0.0503} \\ 
    \cmidrule(r{1mm}){1-1} \cmidrule(l{1mm}r{1mm}){2-4} \cmidrule(l{1mm}r{1mm}){5-7} \cmidrule(l{1mm}r{1mm}){8-10} \cmidrule(l{1mm}){11-13}
    \first{\textsf{\textbf{DeepForm}}}       & \first{0.0096} & \first{0.0136} & \underline{0.0184} & \first{0.0363} & \first{0.0673} & \first{0.1113} & \first{0.0081} & \first{0.0103} & \first{0.0138} & \first{0.0223} & \first{0.0364} & \first{0.0552} \\
    \bottomrule
\end{tabular}
}
\end{subtable}
\vspace{0.9em}
\label{tab:aggregation_strategies}
\end{table*}

\section{Experimental Results}
\label{sec:06_exp_results}
This section summarizes the experimental results from five different settings to highlight the effectiveness of \textsf{\textbf{DeepForm}}.

\subsubsection*{\textbf{Overall Accuracy.}} 
Table \ref{tab:aggregation_strategies} summarizes the overall performance of group recommendation across various recommender algorithms in comparison with baselines on the Baby and Clothing datasets. 
In particular, the Clothing dataset, with its high data sparsity of approximately 99.97\%, exhibits lower accuracy compared to the Baby dataset, which has a sparsity of around 99.88\%. 
Overall, \textsf{\textbf{DeepForm}} consistently demonstrates superior performance. 
Specifically, in AVG and LM, the proposed method always achieves the highest accuracy under all conditions, and in NCF and BC, it outperforms existing studies in most metrics.
Among the baselines, heuristic and clustering methods generally underperform compared to deep graph clustering models. 
In particular, $K$-Means showed the lowest performance in most conditions, while similarity and GMM exhibited slightly better but still comparable performance. 
In contrast, deep graph clustering models exhibits a substantial average performance advantage over conventional group formation approaches, indicating their effectiveness in handling sparse datasets from real-world scenarios. 
Notably, while MGCN achieves strong results under many conditions, it does not exhibit the same level of consistency as \textsf{\textbf{DeepForm}}, primarily because it was not optimized for the RS task.

\begin{figure}[t]
    \centering
    \includegraphics[width=1\linewidth]{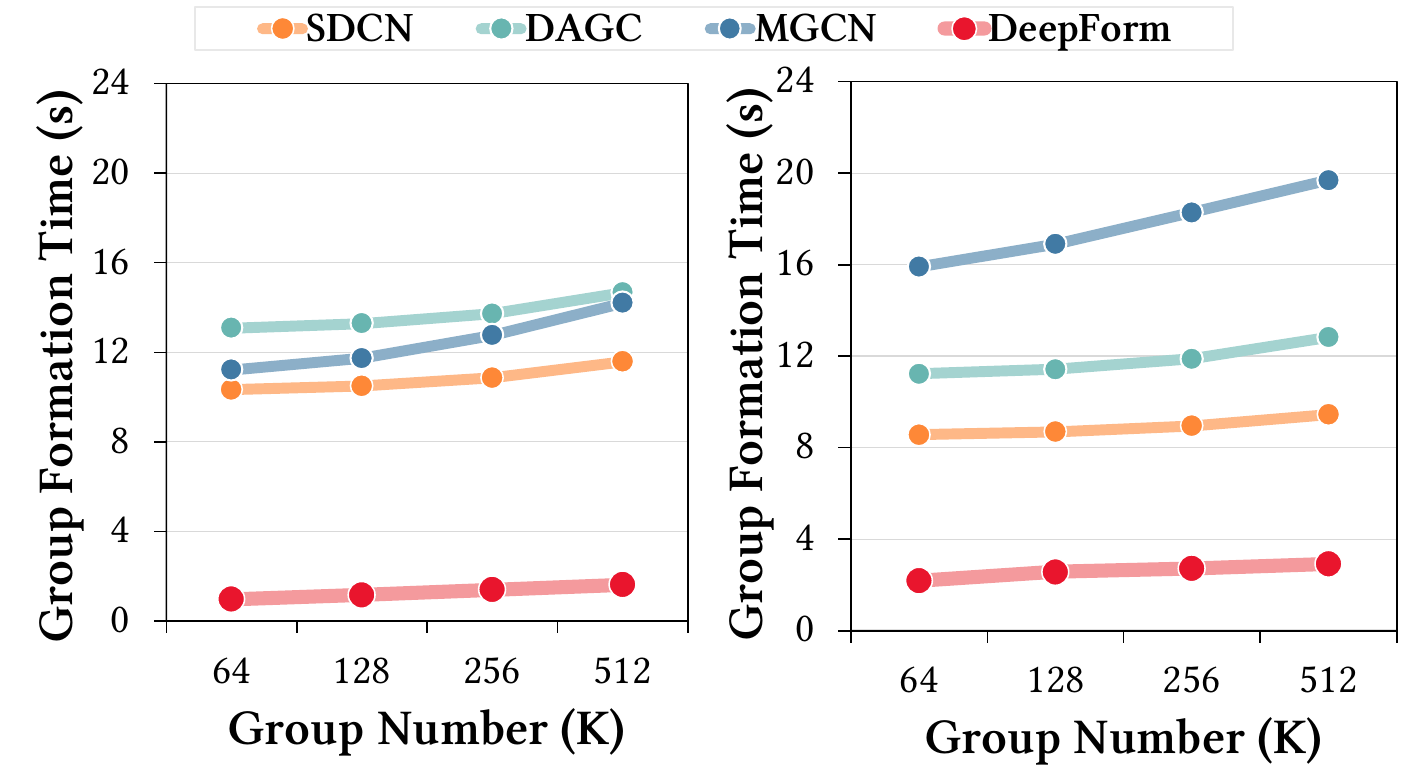}
    \begin{subfigure}{0.4\linewidth}
        \subcaption{\textnormal{Baby: \textsf{\textbf{DeepForm}} is fastest for all groups.}}
    \end{subfigure}%
    \hspace{10pt}
    \begin{subfigure}{0.4\linewidth}
        \subcaption{\textnormal{Clothing: \textsf{\textbf{DeepForm}} is fastest for all groups.}}
    \end{subfigure}%
    \vspace{-2pt}
    \caption{Efficiency comparison of deep graph clustering methods across various numbers of groups.}
    \vspace{0.5em}
\label{fig:efficiency}
\end{figure}

\begin{table}[t]
\vspace{2.5em}
\caption{Ablation study of the effect of removing core components on the Clothing dataset. "Stochastic", "Triplet", and "NCE" denote stochastic cluster learning, triplet loss, and InfoNCE loss, respectively.}
\vspace{-0.2em}
\resizebox{1\columnwidth}{!}{%
\begin{threeparttable}
\renewcommand{\arraystretch}{1}
\centering
\setlength{\tabcolsep}{3pt}
\begin{tabular}{lcccccc}
    \toprule
    \textbf{Ablation study} & \textbf{N@5} & \textbf{N@10} & \textbf{N@20} & \textbf{H@5} & \textbf{H@10} & \textbf{H@20} \\
    \cmidrule(r{0.7mm}){1-1} \cmidrule(l{0.7mm}r{0.7mm}){2-4}  \cmidrule(l{0.7mm}){5-7}
    \textsf{\textbf{DeepForm}}          & \first{0.0114} & \first{0.0136} & \first{0.0138} & \underline{0.0352}         & \first{0.0503} & \first{0.0552} \\
    \cmidrule(r{0.7mm}){1-1} \cmidrule(l{0.7mm}r{0.7mm}){2-4}  \cmidrule(l{0.7mm}){5-7}
    w/o. Stochastic   & 0.0100             & 0.0116             & 0.0132             & 0.0337         & 0.0440             & 0.0518 \\
    w/o. Triplet      & \underline{0.0109} & 0.0125             & 0.0126             & \first{0.0356} & \underline{0.0462} & 0.0492 \\
    w/o. NCE          & \underline{0.0109} & \underline{0.0127} & \underline{0.0136} & 0.0325         & 0.0437             & \underline{0.0522} \\
    \bottomrule
\end{tabular}
\end{threeparttable}
}
\label{tab:ablation}
\vspace{1.4em}
\end{table}

\subsubsection*{\textbf{Efficiency Comparison.}} 
Figure \ref{fig:efficiency} compares the group formation time of high-performing deep graph clustering methods and \textsf{\textbf{DeepForm}} across various numbers of groups. 
Despite the increase in the number of groups, \textsf{\textbf{DeepForm}} consistently maintains a low computational cost. 
In the Baby dataset, it requires only approximately 1 second, which is about 88.03\% faster on average than SDCN, the next fastest method. 
A similar trend is observed in the Clothing dataset, where \textsf{\textbf{DeepForm}} maintains superior efficiency, being approximately 70.90\% faster on average than SDCN.
This performance advantage arises because existing deep graph clustering methods must retrain for each choice of $K$, causing group formation time to increase substantially, whereas \textsf{\textbf{DeepForm}} can dynamically handle various group conditions without additional retraining.
In particular, the minimal cost increase of \textsf{\textbf{DeepForm}} with larger numbers of groups demonstrates both its scalability and adaptability for practical applications in GRSs. 
In summary, by reducing computational overhead by more than four times compared to existing research, \textsf{\textbf{DeepForm}} enables adaptive group recommendation even in large-scale systems where the number of groups changes dynamically.

\begin{figure}[t]
    \centering
    \begin{subfigure}{\linewidth}
        \includegraphics[width=\linewidth]{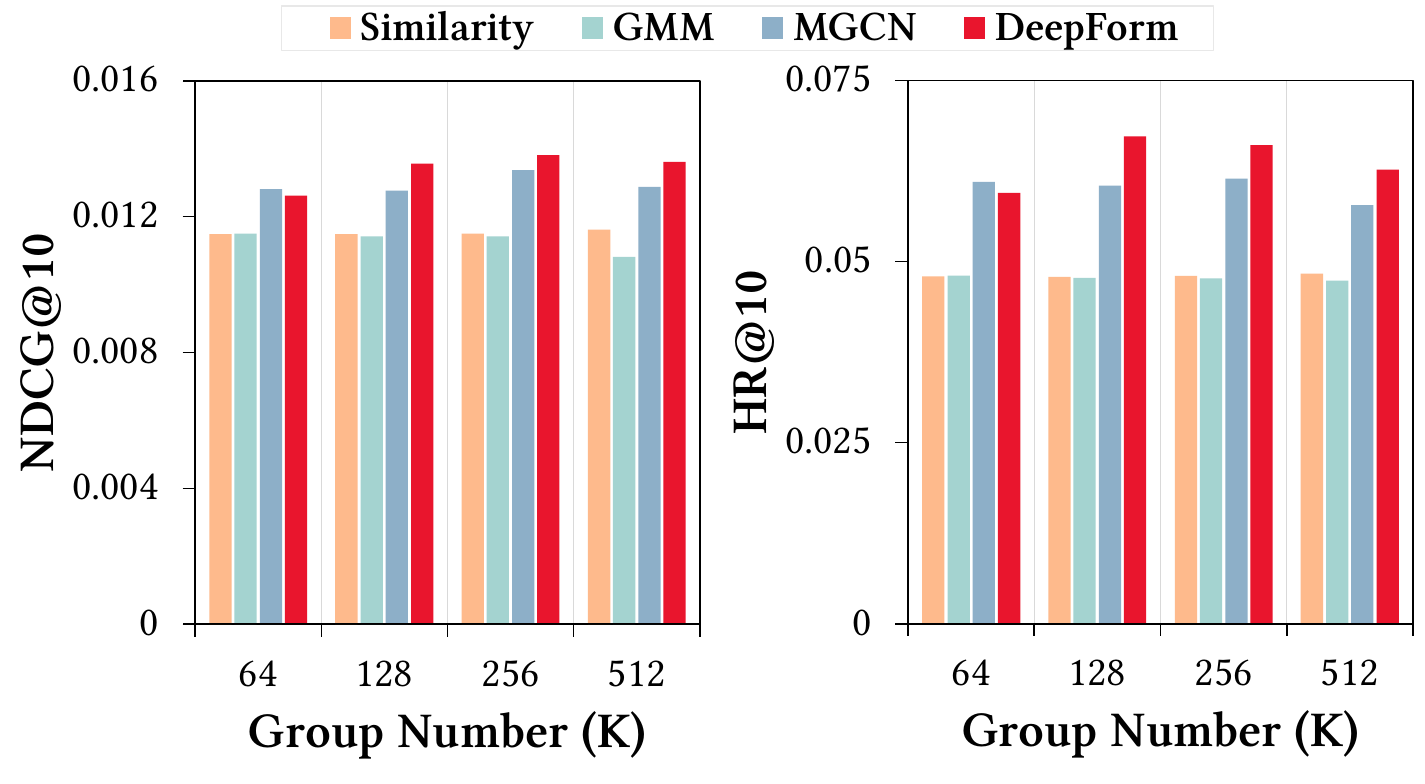}
        \subcaption{\textnormal{Baby: \textsf{\textbf{DeepForm}} achieves the strongest overall performance on most group numbers. MGCN achieves best only at $K=64$.}}
        \vspace{8pt}
    \end{subfigure}
    \begin{subfigure}{\linewidth}
        \includegraphics[width=\linewidth]{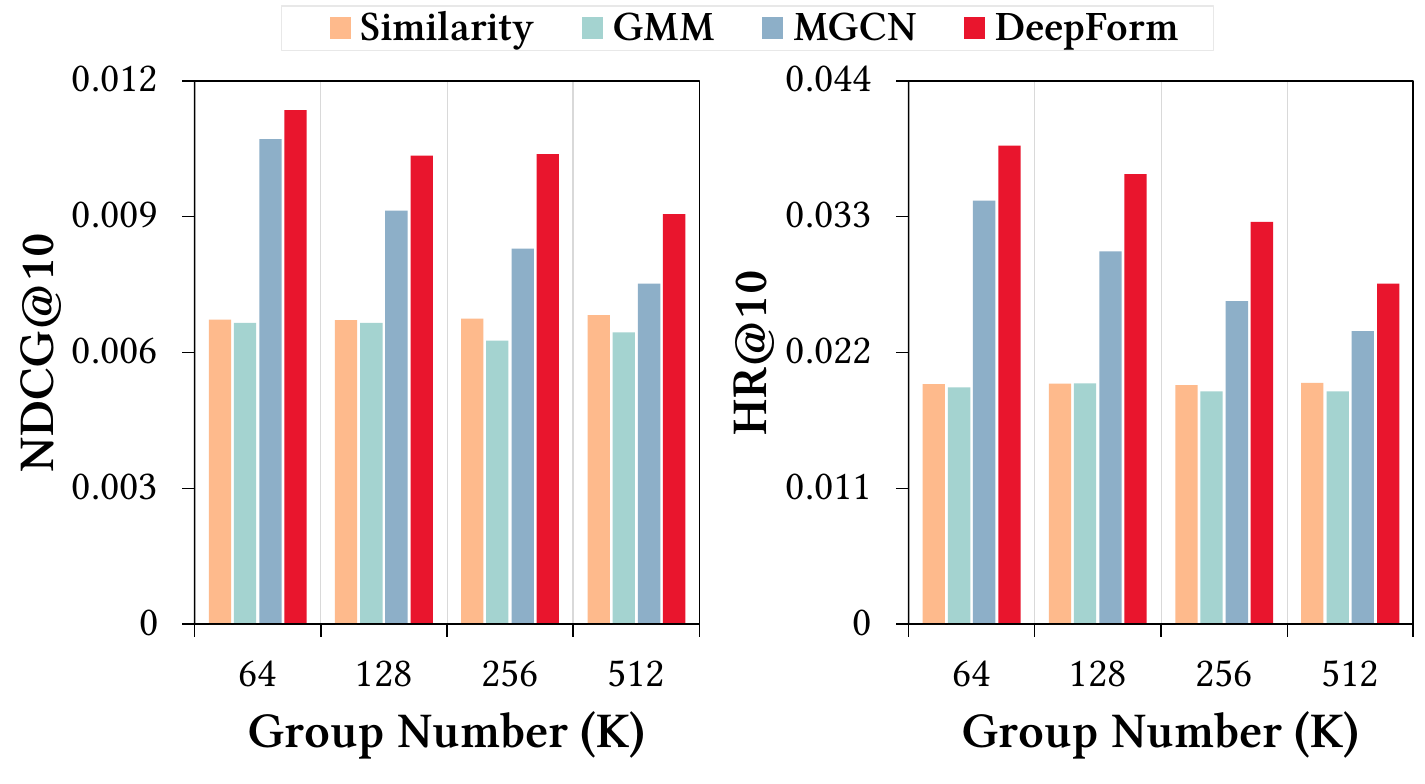}
        \subcaption{\textnormal{Clothing: \textsf{\textbf{DeepForm}} ranks first for all group numbers. Other methods are far behind the proposed method.}}
    \end{subfigure}%
    \caption{Performance comparison across various numbers of groups for representative group formation methods.} 
    \vspace{0.4em}
\label{fig:group_perf}
\end{figure}

\subsubsection*{\textbf{Ablation Study.}}
To verify the contribution of each component in the proposed method, we perform an ablation study as shown in Table \ref{tab:ablation}. 
The experiment is conducted on the Clothing dataset using NCF as the group recommender algorithm, and performance is measured by removing key components from the proposed architecture. 
First, removing stochastic cluster learning results in an average performance drop of approximately 9.05\%, demonstrating the reduced ability to effectively capture diverse user–group structures. 
Second, removing the triplet loss leads to an average drop of approximately 6.51\%, and it shows the lowest performance as top-$k$ recommendations increase, proving its importance in strengthening inter-cluster separation. 
Lastly, removing InfoNCE loss results in the lowest overall performance, reflecting the impact of intra-cluster cohesion. 
In conclusion, the combination of stochastic cluster learning and two contrastive losses provides strong generalization and precise group recommendations for practical GRSs.

\subsubsection*{\textbf{Accuracy by the Number of Groups.}} 
To evaluate the impact of the number of groups, Figure \ref{fig:group_perf} compares representative group formation methods (\emph{i.e.,} Similarity, GMM, and MGCN) including \textsf{\textbf{DeepForm}}. 
In general, the optimal number of groups does not show a clear trend or pattern, as it depends on the characteristics of the datasets and the group formation methods. The degree of performance variation also differs. 
Similarity and GMM show only slight changes across different group numbers, whereas MGCN exhibits substantial fluctuations in accuracy. 
Nevertheless, \textsf{\textbf{DeepForm}} demonstrates consistently strong performance under various conditions with sparse datasets compared to existing research. 
According to the results in Figure \ref{fig:group_perf}, we set $K=128$, which achieves the highest average performance, as the optimal number of groups for all methods in our experiments.

\subsubsection*{\textbf{Effectiveness of Group Formation Methods.}} 
While most recent research on group recommendation relies on predefined groups provided in the dataset, real-world scenarios often require forming groups dynamically. 
To investigate the impact of different group formation strategies, we compare predefined and ad-hoc groups using the high-performing ConsRec on the  CAMRa2011 dataset (see Table \ref{tab:performance_camra}). 
\textsf{\textbf{DeepForm}} achieves approximately 32.57\% higher accuracy than predefined groups in most cases. 
Additionally, \textsf{\textbf{DeepForm}} shows an improvement of about 5.32\% compared to ITR. 
These results indicate the importance of dynamically identifying ad-hoc groups rather than relying solely on predefined groups, and demonstrate that the proposed method is more effective than self-supervised optimal search approaches such as ITR.

\begin{table}[t]
\renewcommand{\arraystretch}{1}
\centering
\caption{Effectiveness comparison of group formation methods on the CAMRa2011 dataset. All recommendations are generated using ConsRec \cite{wu2023consrec}.}
\vspace{-0.5em}
\setlength{\tabcolsep}{3pt}
\resizebox{1\columnwidth}{!}{
\begin{threeparttable}
\begin{tabular}{lcccccc}
    \toprule
    \textbf{Method} & \textbf{N@5} & \textbf{N@10} & \textbf{N@20} & \textbf{H@5} & \textbf{H@10} & \textbf{H@20}        \\ 
    \cmidrule(r{0.7mm}){1-1} \cmidrule(l{0.7mm}r{0.7mm}){2-4}  \cmidrule(l{0.7mm}){5-7}
    Predefined  & 0.4055         & 0.4673         & 0.4964         & 0.6145         & \underline{0.8028}         & \first{0.9166} \\
    ITR$^*$ \cite{liu2024identify}         & \underline{0.6309}         & \underline{0.6714}         & \underline{0.7004}         & \underline{0.6766}         & \underline{0.8028}         & \underline{0.9152} \\
    \textsf{\textbf{DeepForm}}$^*$    & \first{0.7004} & \first{0.7246} & \first{0.7489} & \first{0.7269} & \first{0.8041} & 0.9007 \\ 
    \bottomrule
\end{tabular}
\begin{tablenotes}
\small{
    \item[*] Leverages modified ConsRec \cite{liu2024identify} using only user–item interactions.
}
\end{tablenotes}
\end{threeparttable}
}
\label{tab:performance_camra}
\vspace{1em}
\end{table}

\section{Discussion and Limitations}
\label{sec:07_discussions}
Based on the experimental results, we analyze the impact of the group formation requirements for practical GRSs and discuss potential areas for improvement.

\subsubsection*{\textbf{High-order Graph Structure.}}
Table \ref{tab:aggregation_strategies} clearly shows that the performance gap between conventional group formation approaches and deep graph clustering methods highlights the importance of leveraging graph structure. 
Moreover, user–user relationships often extend beyond direct interactions, making high-order modeling essential for uncovering latent group structures.
To integrate high-order graph information into clustering, we design an architecture that learns user–user relations through a lightweight GCN encoder and complements them with user–item interactions via an AE. 
This design operates efficiently in real-world environments with a large number of items, while jointly optimizing structural and attribute representations to stabilize cluster quality. 
Our experiments demonstrate that this approach consistently improves performance even in extremely sparse scenarios, and the ablation study indicates that incorporating explicit cluster-level contrastive learning further enhances the benefits of high-order modeling.

\subsubsection*{\textbf{Real-time Group Formation.}}
Practical applications in GRSs require optimized inference time for group formation. 
To address this, we introduce stochastic cluster learning, which exposes the model to diverse cluster separations during training, enabling real-time group formation without retraining when the number of groups changes. 
Therefore, as shown in Figure \ref{fig:efficiency}, \textsf{\textbf{DeepForm}} maintains a minimal increase in computational cost as the number of groups grows, while achieving significantly faster group formation compared to existing methods.
This demonstrates that our approach can operate stably even in environments such as streaming services, where resource constraints or operational requirements change frequently.

\subsubsection*{\textbf{Dynamic Determination of the Optimal Number of Groups.}}
As shown in Figure \ref{fig:group_perf}, the optimal number of groups does not follow a consistent pattern.
Furthermore, most existing methods require retraining for each chosen $K$, which is impractical in dynamic environments.
\textsf{\textbf{DeepForm}} demonstrates stable performance across a wide range of $K$ values through stochastic cluster learning and maintains robust representation quality under shifting cluster boundaries through the integration of contrastive learning. 
These results indicate that our approach can maintain recommendation quality without retraining, even in environments where $K$ fluctuates due to resource changes or evolving user behaviors.
This robustness stems from exposing the model to a diverse range of clustering granularities during training, which mitigates sensitivity to any specific choice of $K$.

\subsubsection*{\textbf{Limitations and Future Work.}}
First, while stochastic cluster learning expands the search space and enables generalization across $K$ values, training costs may increase substantially when covering a wide range of cases. 
In this regard, heuristic strategies such as mixed-precision training, partial centroid updates, or progressive learning may help reduce this overhead.
Second, although experiments demonstrate that our method works effectively in sparse data environments, the representations still rely on user–item interactions, making it difficult to directly address the cold-start problem. 
Potential solutions include incorporating external metadata, adding auxiliary encoders, or adopting meta-learning approaches.
Lastly, although our method is designed for computational efficiency in practical group formation, handling extremely large-scale networks (\emph{e.g.,} hundreds of millions of users and items) in real-time may require additional strategies such as distributed processing or model compression.

\section{Conclusion}
\label{sec:08_conclusion}

We identify three practical challenges for group formation in GRSs: (1) incorporating high-order graph structures, (2) enabling real-time group formation, and (3) dynamically adjusting the number of groups.
We present \textsf{\textbf{DeepForm}}, a framework that addresses all three.
\textsf{\textbf{DeepForm}} uses a propagation-only deep graph clustering architecture to encode high-order structure from user–item interactions, 
trains with stochastic cluster learning for real-time group formation, and leverages contrastive learning for robustness under changing conditions in real-word environments of GRSs.
Across diverse scenarios, \textsf{\textbf{DeepForm}} delivers consistent gains in accuracy and efficiency. These results highlight the central role of group formation and delineate the requirements for real-world deployment.


\balance
\bibliographystyle{ACM-Reference-Format}
\bibliography{10_References}


\begin{thebibliography}{32}


\ifx \showCODEN    \undefined \def \showCODEN     #1{\unskip}     \fi
\ifx \showISBNx    \undefined \def \showISBNx     #1{\unskip}     \fi
\ifx \showISBNxiii \undefined \def \showISBNxiii  #1{\unskip}     \fi
\ifx \showISSN     \undefined \def \showISSN      #1{\unskip}     \fi
\ifx \showLCCN     \undefined \def \showLCCN      #1{\unskip}     \fi
\ifx \shownote     \undefined \def \shownote      #1{#1}          \fi
\ifx \showarticletitle \undefined \def \showarticletitle #1{#1}   \fi
\ifx \showURL      \undefined \def \showURL       {\relax}        \fi
\providecommand\bibfield[2]{#2}
\providecommand\bibinfo[2]{#2}
\providecommand\natexlab[1]{#1}
\providecommand\showeprint[2][]{arXiv:#2}

\bibitem[Alves et~al\mbox{.}(2022)]%
        {alves2022grouplanner}
\bibfield{author}{\bibinfo{person}{Patr{\'\i}cia Alves}, \bibinfo{person}{Domingos Gomes}, \bibinfo{person}{Catarina Rodrigues}, \bibinfo{person}{Jo{\~a}o Carneiro}, \bibinfo{person}{Paulo Novais}, {and} \bibinfo{person}{Goreti Marreiros}.} \bibinfo{year}{2022}\natexlab{}.
\newblock \showarticletitle{Grouplanner: a group recommender system for tourism with multi-agent microservices}. In \bibinfo{booktitle}{\emph{International Conference on Practical Applications of Agents and Multi-Agent Systems}}. Springer, \bibinfo{pages}{454--460}.
\newblock


\bibitem[Anand and Maurya(2025)]%
        {anand2025survey}
\bibfield{author}{\bibinfo{person}{Vineeta Anand} {and} \bibinfo{person}{Ashish~Kumar Maurya}.} \bibinfo{year}{2025}\natexlab{}.
\newblock \showarticletitle{A survey on recommender systems using graph neural network}.
\newblock \bibinfo{journal}{\emph{ACM Transactions on Information Systems}} \bibinfo{volume}{43}, \bibinfo{number}{1} (\bibinfo{year}{2025}), \bibinfo{pages}{1--49}.
\newblock


\bibitem[Baltrunas et~al\mbox{.}(2010)]%
        {baltrunas2010group}
\bibfield{author}{\bibinfo{person}{Linas Baltrunas}, \bibinfo{person}{Tadas Makcinskas}, {and} \bibinfo{person}{Francesco Ricci}.} \bibinfo{year}{2010}\natexlab{}.
\newblock \showarticletitle{Group recommendations with rank aggregation and collaborative filtering}. In \bibinfo{booktitle}{\emph{Proceedings of the fourth ACM conference on Recommender systems}}. \bibinfo{pages}{119--126}.
\newblock


\bibitem[Bo et~al\mbox{.}(2020)]%
        {bo2020structural}
\bibfield{author}{\bibinfo{person}{Deyu Bo}, \bibinfo{person}{Xiao Wang}, \bibinfo{person}{Chuan Shi}, \bibinfo{person}{Meiqi Zhu}, \bibinfo{person}{Emiao Lu}, {and} \bibinfo{person}{Peng Cui}.} \bibinfo{year}{2020}\natexlab{}.
\newblock \showarticletitle{Structural deep clustering network}. In \bibinfo{booktitle}{\emph{Proceedings of the web conference 2020}}. \bibinfo{pages}{1400--1410}.
\newblock


\bibitem[Cao et~al\mbox{.}(2018)]%
        {cao2018attentive}
\bibfield{author}{\bibinfo{person}{Da Cao}, \bibinfo{person}{Xiangnan He}, \bibinfo{person}{Lianhai Miao}, \bibinfo{person}{Yahui An}, \bibinfo{person}{Chao Yang}, {and} \bibinfo{person}{Richang Hong}.} \bibinfo{year}{2018}\natexlab{}.
\newblock \showarticletitle{Attentive group recommendation}. In \bibinfo{booktitle}{\emph{The 41st International ACM SIGIR conference on research \& development in information retrieval}}. \bibinfo{pages}{645--654}.
\newblock


\bibitem[Cao et~al\mbox{.}(2019)]%
        {cao2019social}
\bibfield{author}{\bibinfo{person}{Da Cao}, \bibinfo{person}{Xiangnan He}, \bibinfo{person}{Lianhai Miao}, \bibinfo{person}{Guangyi Xiao}, \bibinfo{person}{Hao Chen}, {and} \bibinfo{person}{Jiao Xu}.} \bibinfo{year}{2019}\natexlab{}.
\newblock \showarticletitle{Social-enhanced attentive group recommendation}.
\newblock \bibinfo{journal}{\emph{IEEE Transactions on Knowledge and Data Engineering}} \bibinfo{volume}{33}, \bibinfo{number}{3} (\bibinfo{year}{2019}), \bibinfo{pages}{1195--1209}.
\newblock


\bibitem[Ceh-Varela et~al\mbox{.}(2022)]%
        {ceh2022performance}
\bibfield{author}{\bibinfo{person}{Edgar Ceh-Varela}, \bibinfo{person}{Huiping Cao}, {and} \bibinfo{person}{Hady~W Lauw}.} \bibinfo{year}{2022}\natexlab{}.
\newblock \showarticletitle{Performance evaluation of aggregation-based group recommender systems for ephemeral groups}.
\newblock \bibinfo{journal}{\emph{ACM Transactions on Intelligent Systems and Technology (TIST)}} \bibinfo{volume}{13}, \bibinfo{number}{6} (\bibinfo{year}{2022}), \bibinfo{pages}{1--26}.
\newblock


\bibitem[Chen et~al\mbox{.}(2018)]%
        {chen2018hybrid}
\bibfield{author}{\bibinfo{person}{Rui Chen}, \bibinfo{person}{Qingyi Hua}, \bibinfo{person}{Quanli Gao}, {and} \bibinfo{person}{Ying Xing}.} \bibinfo{year}{2018}\natexlab{}.
\newblock \showarticletitle{A hybrid recommender system for Gaussian mixture model and enhanced social matrix factorization technology based on multiple interests}.
\newblock \bibinfo{journal}{\emph{Mathematical Problems in Engineering}} \bibinfo{volume}{2018}, \bibinfo{number}{1} (\bibinfo{year}{2018}), \bibinfo{pages}{9109647}.
\newblock


\bibitem[Chen et~al\mbox{.}(2022)]%
        {chen2022thinking}
\bibfield{author}{\bibinfo{person}{Tong Chen}, \bibinfo{person}{Hongzhi Yin}, \bibinfo{person}{Jing Long}, \bibinfo{person}{Quoc Viet~Hung Nguyen}, \bibinfo{person}{Yang Wang}, {and} \bibinfo{person}{Meng Wang}.} \bibinfo{year}{2022}\natexlab{}.
\newblock \showarticletitle{Thinking inside the box: learning hypercube representations for group recommendation}. In \bibinfo{booktitle}{\emph{Proceedings of the 45th International ACM SIGIR Conference on Research and Development in Information Retrieval}}. \bibinfo{pages}{1664--1673}.
\newblock


\bibitem[Dong and Shen(2018)]%
        {dong2018triplet}
\bibfield{author}{\bibinfo{person}{Xingping Dong} {and} \bibinfo{person}{Jianbing Shen}.} \bibinfo{year}{2018}\natexlab{}.
\newblock \showarticletitle{Triplet loss in siamese network for object tracking}. In \bibinfo{booktitle}{\emph{Proceedings of the European conference on computer vision (ECCV)}}. \bibinfo{pages}{459--474}.
\newblock


\bibitem[Ghazarian and Nematbakhsh(2015)]%
        {ghazarian2015enhancing}
\bibfield{author}{\bibinfo{person}{Sarik Ghazarian} {and} \bibinfo{person}{Mohammad~Ali Nematbakhsh}.} \bibinfo{year}{2015}\natexlab{}.
\newblock \showarticletitle{Enhancing memory-based collaborative filtering for group recommender systems}.
\newblock \bibinfo{journal}{\emph{Expert systems with applications}} \bibinfo{volume}{42}, \bibinfo{number}{7} (\bibinfo{year}{2015}), \bibinfo{pages}{3801--3812}.
\newblock


\bibitem[He et~al\mbox{.}(2020)]%
        {he2020lightgcn}
\bibfield{author}{\bibinfo{person}{Xiangnan He}, \bibinfo{person}{Kuan Deng}, \bibinfo{person}{Xiang Wang}, \bibinfo{person}{Yan Li}, \bibinfo{person}{Yongdong Zhang}, {and} \bibinfo{person}{Meng Wang}.} \bibinfo{year}{2020}\natexlab{}.
\newblock \showarticletitle{Lightgcn: Simplifying and powering graph convolution network for recommendation}. In \bibinfo{booktitle}{\emph{Proceedings of the 43rd International ACM SIGIR conference on research and development in Information Retrieval}}. \bibinfo{pages}{639--648}.
\newblock


\bibitem[Htun et~al\mbox{.}(2021)]%
        {htun2021perception}
\bibfield{author}{\bibinfo{person}{Nyi~Nyi Htun}, \bibinfo{person}{Elisa Lecluse}, {and} \bibinfo{person}{Katrien Verbert}.} \bibinfo{year}{2021}\natexlab{}.
\newblock \showarticletitle{Perception of fairness in group music recommender systems}. In \bibinfo{booktitle}{\emph{Proceedings of the 26th International Conference on Intelligent User Interfaces}}. \bibinfo{pages}{302--306}.
\newblock


\bibitem[Krouska et~al\mbox{.}(2023)]%
        {krouska2023novel}
\bibfield{author}{\bibinfo{person}{Akrivi Krouska}, \bibinfo{person}{Christos Troussas}, {and} \bibinfo{person}{Cleo Sgouropoulou}.} \bibinfo{year}{2023}\natexlab{}.
\newblock \showarticletitle{A novel group recommender system for domain-independent decision support customizing a grouping genetic algorithm}.
\newblock \bibinfo{journal}{\emph{User Modeling and User-Adapted Interaction}} \bibinfo{volume}{33}, \bibinfo{number}{5} (\bibinfo{year}{2023}), \bibinfo{pages}{1113--1140}.
\newblock


\bibitem[Kumar and Chowdary(2022)]%
        {kumar2022ophaencoder}
\bibfield{author}{\bibinfo{person}{Chintoo Kumar} {and} \bibinfo{person}{C~Ravindranath Chowdary}.} \bibinfo{year}{2022}\natexlab{}.
\newblock \showarticletitle{OPHAencoder: An unsupervised approach to identify groups in group recommendations}.
\newblock \bibinfo{journal}{\emph{Computing}} \bibinfo{volume}{104}, \bibinfo{number}{12} (\bibinfo{year}{2022}), \bibinfo{pages}{2635--2657}.
\newblock


\bibitem[Leiber et~al\mbox{.}(2024)]%
        {leiber2024dying}
\bibfield{author}{\bibinfo{person}{Collin Leiber}, \bibinfo{person}{Niklas Strau{\ss}}, \bibinfo{person}{Matthias Schubert}, {and} \bibinfo{person}{Thomas Seidl}.} \bibinfo{year}{2024}\natexlab{}.
\newblock \showarticletitle{Dying Clusters Is All You Need-Deep Clustering With an Unknown Number of Clusters}. In \bibinfo{booktitle}{\emph{2024 IEEE International Conference on Data Mining Workshops (ICDMW)}}. IEEE, \bibinfo{pages}{726--733}.
\newblock


\bibitem[Li et~al\mbox{.}(2024)]%
        {li2024multi}
\bibfield{author}{\bibinfo{person}{Xiulai Li}, \bibinfo{person}{Wei Wu}, \bibinfo{person}{Bin Zhang}, {and} \bibinfo{person}{Xin Peng}.} \bibinfo{year}{2024}\natexlab{}.
\newblock \showarticletitle{Multi-scale Graph Clustering Network}.
\newblock \bibinfo{journal}{\emph{Information Sciences}} (\bibinfo{year}{2024}), \bibinfo{pages}{121023}.
\newblock


\bibitem[Liu et~al\mbox{.}(2022)]%
        {liu2022survey}
\bibfield{author}{\bibinfo{person}{Yue Liu}, \bibinfo{person}{Jun Xia}, \bibinfo{person}{Sihang Zhou}, \bibinfo{person}{Xihong Yang}, \bibinfo{person}{Ke Liang}, \bibinfo{person}{Chenchen Fan}, \bibinfo{person}{Yan Zhuang}, \bibinfo{person}{Stan~Z Li}, \bibinfo{person}{Xinwang Liu}, {and} \bibinfo{person}{Kunlun He}.} \bibinfo{year}{2022}\natexlab{}.
\newblock \showarticletitle{A survey of deep graph clustering: Taxonomy, challenge, application, and open resource}.
\newblock \bibinfo{journal}{\emph{arXiv preprint arXiv:2211.12875}} (\bibinfo{year}{2022}).
\newblock


\bibitem[Liu et~al\mbox{.}(2024)]%
        {liu2024identify}
\bibfield{author}{\bibinfo{person}{Yue Liu}, \bibinfo{person}{Shihao Zhu}, \bibinfo{person}{Tianyuan Yang}, \bibinfo{person}{Jian Ma}, {and} \bibinfo{person}{Wenliang Zhong}.} \bibinfo{year}{2024}\natexlab{}.
\newblock \showarticletitle{Identify then recommend: Towards unsupervised group recommendation}.
\newblock \bibinfo{journal}{\emph{Advances in Neural Information Processing Systems}}  \bibinfo{volume}{37} (\bibinfo{year}{2024}), \bibinfo{pages}{96101--96126}.
\newblock


\bibitem[Nozari and Koohi(2020)]%
        {nozari2020novel}
\bibfield{author}{\bibinfo{person}{Reza~Barzegar Nozari} {and} \bibinfo{person}{Hamidreza Koohi}.} \bibinfo{year}{2020}\natexlab{}.
\newblock \showarticletitle{A novel group recommender system based on members’ influence and leader impact}.
\newblock \bibinfo{journal}{\emph{Knowledge-Based Systems}}  \bibinfo{volume}{205} (\bibinfo{year}{2020}), \bibinfo{pages}{106296}.
\newblock


\bibitem[O’connor et~al\mbox{.}(2001)]%
        {o2001polylens}
\bibfield{author}{\bibinfo{person}{Mark O’connor}, \bibinfo{person}{Dan Cosley}, \bibinfo{person}{Joseph~A Konstan}, {and} \bibinfo{person}{John Riedl}.} \bibinfo{year}{2001}\natexlab{}.
\newblock \showarticletitle{PolyLens: A recommender system for groups of users}. In \bibinfo{booktitle}{\emph{ECSCW 2001: Proceedings of the Seventh European conference on computer supported cooperative work 16--20 September 2001, Bonn, Germany}}. Springer, \bibinfo{pages}{199--218}.
\newblock


\bibitem[Peng et~al\mbox{.}(2022)]%
        {peng2022deep}
\bibfield{author}{\bibinfo{person}{Zhihao Peng}, \bibinfo{person}{Hui Liu}, \bibinfo{person}{Yuheng Jia}, {and} \bibinfo{person}{Junhui Hou}.} \bibinfo{year}{2022}\natexlab{}.
\newblock \showarticletitle{Deep attention-guided graph clustering with dual self-supervision}.
\newblock \bibinfo{journal}{\emph{IEEE Transactions on Circuits and Systems for Video Technology}} \bibinfo{volume}{33}, \bibinfo{number}{7} (\bibinfo{year}{2022}), \bibinfo{pages}{3296--3307}.
\newblock


\bibitem[Pujahari and Padmanabhan(2015)]%
        {pujahari2015group}
\bibfield{author}{\bibinfo{person}{Abinash Pujahari} {and} \bibinfo{person}{Vineet Padmanabhan}.} \bibinfo{year}{2015}\natexlab{}.
\newblock \showarticletitle{Group recommender systems: Combining user-user and item-item collaborative filtering techniques}. In \bibinfo{booktitle}{\emph{2015 International conference on information technology (ICIT)}}. IEEE, \bibinfo{pages}{148--152}.
\newblock


\bibitem[Sankar et~al\mbox{.}(2020)]%
        {sankar2020groupim}
\bibfield{author}{\bibinfo{person}{Aravind Sankar}, \bibinfo{person}{Yanhong Wu}, \bibinfo{person}{Yuhang Wu}, \bibinfo{person}{Wei Zhang}, \bibinfo{person}{Hao Yang}, {and} \bibinfo{person}{Hari Sundaram}.} \bibinfo{year}{2020}\natexlab{}.
\newblock \showarticletitle{Groupim: A mutual information maximization framework for neural group recommendation}. In \bibinfo{booktitle}{\emph{Proceedings of the 43rd International ACM SIGIR conference on research and development in Information Retrieval}}. \bibinfo{pages}{1279--1288}.
\newblock


\bibitem[Seo et~al\mbox{.}(2021)]%
        {seo2021group}
\bibfield{author}{\bibinfo{person}{Young-Duk Seo}, \bibinfo{person}{Young-Gab Kim}, \bibinfo{person}{Euijong Lee}, {and} \bibinfo{person}{Hyungjin Kim}.} \bibinfo{year}{2021}\natexlab{}.
\newblock \showarticletitle{Group recommender system based on genre preference focusing on reducing the clustering cost}.
\newblock \bibinfo{journal}{\emph{Expert Systems with Applications}}  \bibinfo{volume}{183} (\bibinfo{year}{2021}), \bibinfo{pages}{115396}.
\newblock


\bibitem[Stratigi et~al\mbox{.}(2022)]%
        {stratigi2022sequential}
\bibfield{author}{\bibinfo{person}{Maria Stratigi}, \bibinfo{person}{Evaggelia Pitoura}, \bibinfo{person}{Jyrki Nummenmaa}, {and} \bibinfo{person}{Kostas Stefanidis}.} \bibinfo{year}{2022}\natexlab{}.
\newblock \showarticletitle{Sequential group recommendations based on satisfaction and disagreement scores}.
\newblock \bibinfo{journal}{\emph{Journal of Intelligent Information Systems}} \bibinfo{volume}{58}, \bibinfo{number}{2} (\bibinfo{year}{2022}), \bibinfo{pages}{227--254}.
\newblock


\bibitem[Wu et~al\mbox{.}(2022)]%
        {wu2022graph}
\bibfield{author}{\bibinfo{person}{Shiwen Wu}, \bibinfo{person}{Fei Sun}, \bibinfo{person}{Wentao Zhang}, \bibinfo{person}{Xu Xie}, {and} \bibinfo{person}{Bin Cui}.} \bibinfo{year}{2022}\natexlab{}.
\newblock \showarticletitle{Graph neural networks in recommender systems: a survey}.
\newblock \bibinfo{journal}{\emph{Comput. Surveys}} \bibinfo{volume}{55}, \bibinfo{number}{5} (\bibinfo{year}{2022}), \bibinfo{pages}{1--37}.
\newblock


\bibitem[Wu et~al\mbox{.}(2023)]%
        {wu2023consrec}
\bibfield{author}{\bibinfo{person}{Xixi Wu}, \bibinfo{person}{Yun Xiong}, \bibinfo{person}{Yao Zhang}, \bibinfo{person}{Yizhu Jiao}, \bibinfo{person}{Jiawei Zhang}, \bibinfo{person}{Yangyong Zhu}, {and} \bibinfo{person}{Philip~S Yu}.} \bibinfo{year}{2023}\natexlab{}.
\newblock \showarticletitle{Consrec: Learning consensus behind interactions for group recommendation}. In \bibinfo{booktitle}{\emph{Proceedings of the acm web conference 2023}}. \bibinfo{pages}{240--250}.
\newblock


\bibitem[Xu et~al\mbox{.}(2024)]%
        {xu2024aligngroup}
\bibfield{author}{\bibinfo{person}{Jinfeng Xu}, \bibinfo{person}{Zheyu Chen}, \bibinfo{person}{Jinze Li}, \bibinfo{person}{Shuo Yang}, \bibinfo{person}{Hewei Wang}, {and} \bibinfo{person}{Edith~CH Ngai}.} \bibinfo{year}{2024}\natexlab{}.
\newblock \showarticletitle{Aligngroup: Learning and aligning group consensus with member preferences for group recommendation}. In \bibinfo{booktitle}{\emph{Proceedings of the 33rd ACM International Conference on Information and Knowledge Management}}. \bibinfo{pages}{2682--2691}.
\newblock


\bibitem[Yalcin et~al\mbox{.}(2021)]%
        {yalcin2021entropy}
\bibfield{author}{\bibinfo{person}{Emre Yalcin}, \bibinfo{person}{Firat Ismailoglu}, {and} \bibinfo{person}{Alper Bilge}.} \bibinfo{year}{2021}\natexlab{}.
\newblock \showarticletitle{An entropy empowered hybridized aggregation technique for group recommender systems}.
\newblock \bibinfo{journal}{\emph{Expert Systems with Applications}}  \bibinfo{volume}{166} (\bibinfo{year}{2021}), \bibinfo{pages}{114111}.
\newblock


\bibitem[Yannam et~al\mbox{.}(2023)]%
        {yannam2023enhancing}
\bibfield{author}{\bibinfo{person}{V~Ramanjaneyulu Yannam}, \bibinfo{person}{Jitendra Kumar}, \bibinfo{person}{Korra~Sathya Babu}, {and} \bibinfo{person}{Bidyut~Kumar Patra}.} \bibinfo{year}{2023}\natexlab{}.
\newblock \showarticletitle{Enhancing the accuracy of group recommendation using slope one}.
\newblock \bibinfo{journal}{\emph{The journal of supercomputing}} \bibinfo{volume}{79}, \bibinfo{number}{1} (\bibinfo{year}{2023}), \bibinfo{pages}{499--540}.
\newblock


\bibitem[Zhu and Wang(2020)]%
        {zhu2020context}
\bibfield{author}{\bibinfo{person}{Qiliang Zhu} {and} \bibinfo{person}{Lei Wang}.} \bibinfo{year}{2020}\natexlab{}.
\newblock \showarticletitle{Context-aware restaurant recommendation for group of people}. In \bibinfo{booktitle}{\emph{2020 IEEE World Congress on Services (SERVICES)}}. IEEE, \bibinfo{pages}{51--54}.
\newblock


\end{thebibliography}

\end{document}